\documentclass{article}

\usepackage{amsmath,amssymb}
\usepackage{graphicx}
\usepackage{caption}
\usepackage{subcaption}

\usepackage{booktabs}
\usepackage{float}
\usepackage{lipsum}
\usepackage{afterpage}
\usepackage{multirow}
\usepackage[table]{xcolor}
\usepackage{colortbl}
\usepackage{hhline}
\usepackage{color, xcolor}
\usepackage{adjustbox}
\usepackage{tabularx}
\usepackage{siunitx}
\usepackage{pifont}
\usepackage{tikz}
\usepackage{tabularx}
\usepackage{colortbl}
\usepackage{makecell}
\usepackage{enumitem}
\usepackage{wasysym}
\usepackage{pifont}

\usepackage{float}
\usepackage{wrapfig}

\newcommand{\blank}{\textunderscore \textunderscore \ }

\newcommand{\cmark}{\ding{51}}
\newcommand{\xmark}{\ding{55}}

\definecolor{brightpink}{rgb}{1.0, 0.0, 0.5}
\definecolor{ao(english)}{rgb}{0.0, 0.5, 0.0}
\definecolor{blue(ncs)}{rgb}{0.0, 0.53, 0.74}
\definecolor{babypink}{rgb}{0.96, 0.76, 0.76}
\definecolor{successColor}{rgb}{0.74, 0.83, 0.9}
\definecolor{blizzardblue}{rgb}{0.67, 0.9, 0.93}

\newcommand{\myPink}[1]{\textcolor{brightpink}{#1}}

\usepackage[preprint]{corl_2025} 

\title{GLOVER++: Unleashing the Potential of Affordance Learning from Human Behaviors for Robotic Manipulation}

\author{
 Teli Ma$^{1,\dagger}$, Jia Zheng$^{1,\dagger}$, Zifan Wang$^1$, Ziyao Gao$^1$, Jiaming Zhou$^1$, Junwei Liang$^{1,2,*}$\\ 
  $^1$ HKUST (GZ) \quad $^2$ HKUST\\
  $^{\dagger}$ Equal contribution. $^*$ Corresponding author.\\ \\
  \href{https://teleema.github.io/projects/GLOVER++/}{teleema.github.io/projects/GLOVER++} \\
}

\begin{document}
\maketitle


\begin{abstract}
Learning manipulation skills from human demonstration videos offers a promising path toward generalizable and interpretable robotic intelligence—particularly through the lens of \textit{actionable affordances}. 
However, transferring such knowledge remains challenging due to:  1) a lack of large-scale datasets with precise affordance annotations, and 2) insufficient exploration of affordances in diverse manipulation contexts. 
To address these gaps, we introduce \textbf{HOVA-500K}, a large-scale, affordance-annotated dataset comprising 500,000 images across 1,726 object categories and 675 actions. We also release a standardized benchmarking suite for multi-modal affordance reasoning.
Built upon HOVA-500K, we present \textbf{GLOVER++}, a \textit{global-to-local} affordance training framework that effectively transfers actionable affordance knowledge from human demonstrations to downstream open-vocabulary reasoning tasks.
GLOVER++ achieves state-of-the-art results on the HOVA-500K benchmark and demonstrates strong generalization across diverse downstream robotic manipulation tasks. 
By explicitly modeling actionable affordances, GLOVER++ facilitates robust transfer across scenes, modalities, and tasks.
We hope that HOVA-500K and the GLOVER++ framework will serve as valuable resources for bridging the gap between human demonstrations and robotic manipulation capabilities.
\end{abstract}

\begin{figure*}[h!]
\vspace{-10pt}
\centering
\includegraphics[width=1.0\linewidth,trim={0cm 0cm 0cm 0cm}]{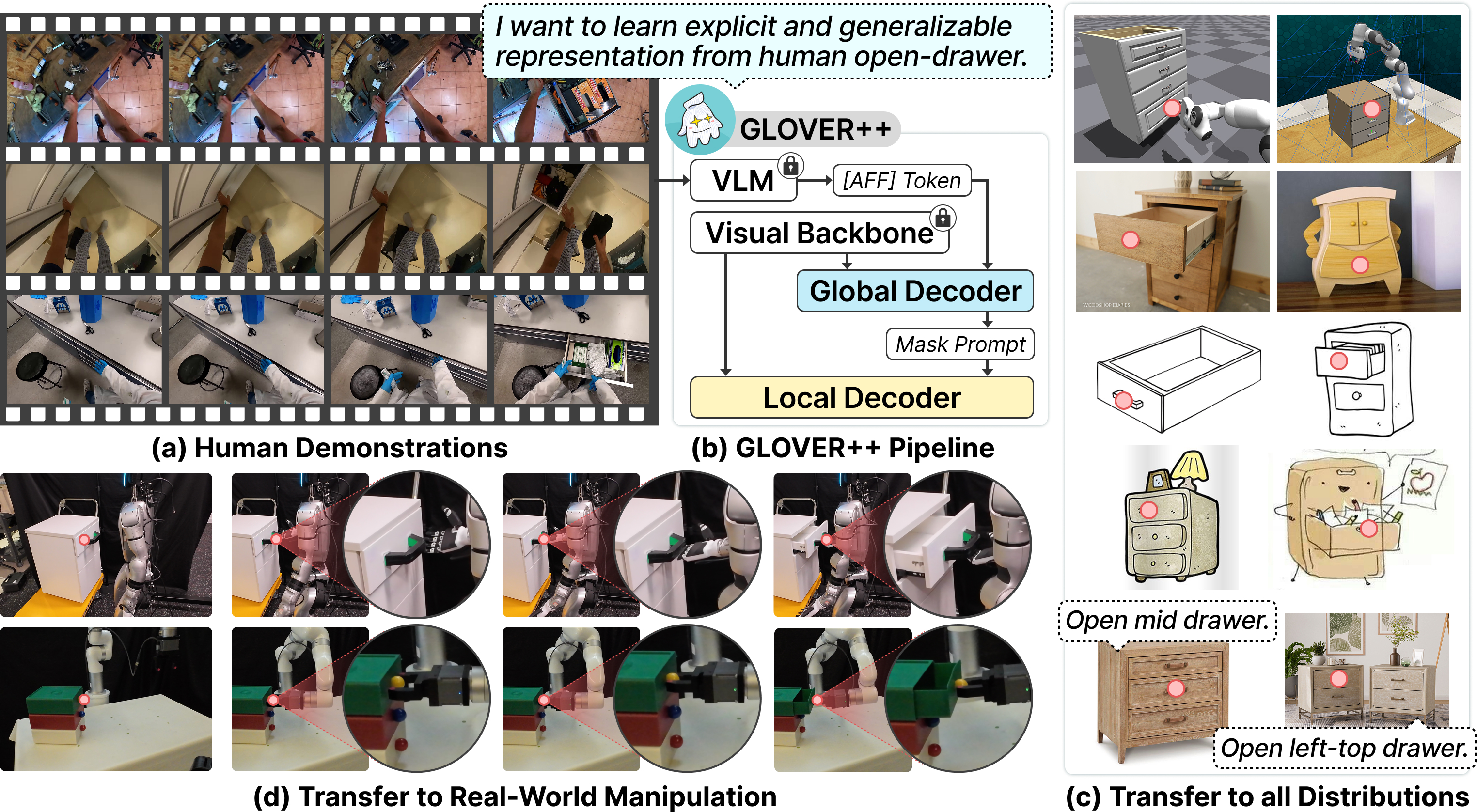}
\caption{(\textbf{a}) GLOVER++ aims to learn generalizable affordance representation from human behaviors (\textit{e.g.} \textit{open drawer)}. (\textbf{b}) The training pipeline of GLOVER++. We adopt a \textit{global-to-local} decoding policy to balance global semantic decoding and local affordance decoding. (\textbf{c}) GLOVER++ is capable of transferring affordable knowledge to all kinds of distributions (\textit{simulation, sketch, cartoon} \textit{etc.}) in an open-vocabulary manner. It also presents strong spatial reasoning ability as shown in the bottom line. (\textbf{d}) By lifting inferred affordable points into 3D space, GLOVER++ provides perceptive awareness for real-world manipulation tasks.
(Red dots represent affordable points.)} 
\label{fig:pipeline}
\vspace{-20pt}
\end{figure*}

\section{Introduction}
Humans can naturally manipulate objects by following language instructions—distinguishing object types, locating them, and choosing affordable parts based on the task in a generalizable way. Hence, images and videos that depict human-object interaction and manipulation are prevalent~\cite{deng2009imagenet, kay2017kinetics, grauman2022ego4d, zhou2023adafocus, zhou2023twinformer}, as such scenarios are highly common in our daily life and easily collectible. 
What can these data do for robotic manipulation? An intrinsic idea is absorbing the potential knowledge embodied in daily human behaviors and transferring it to facilitate robotic manipulation. However, how such knowledge can be learned and transferred remains unclear. 

Some previous works~\cite{lin2024spawnnet, nair2022r3m, xiao2022mvp, zhou2024mitigating} focus on the policy of pretraining in human videos and finetuning in downstream robotic tasks, which reveals limited generalizability and lacks robustness to scene changes. Instead, recent works have paid attention to much more explicit and generalizable representations like \textbf{affordance}~\cite{roboabc, kuang2024ram, affordancellm, ma2024glover}, which refers to relational properties, or potentials for interaction between the environment and the animal as introduced by James J. Gibson~\cite{gibson1977theory}. 
The affordance embodies actionable human knowledge, reflecting the possibility of \textit{where} and \textit{how} to act.
Previous affordance-based methods can be mainly categorized as 3D radiance field modeling~\cite{zheng2024gaussiangrasper, lerftogo, huang2023voxposer, geff}, object retrieval~\cite{kuang2024ram, roboabc}, and vision-language model (VLM) reasoning~\cite{affordancellm, ma2024glover}. However, existing methods have yet to adequately address how to distill actionable affordance knowledge from rich human videos, and how to demonstrate the effective transfer as an explicit representation for a variety of manipulation tasks.

To that end, we introduce \textbf{HOVA-500K}, a large-scale affordance-annotated dataset constructed from existing human videos and images.
The HOVA-500K comprises 500,000 meticulously annotated images spanning 1,726 object categories and 675 action categories, creating a comprehensive taxonomy of human-object interactions. 
HOVA-500K offers three key advantages: 
1) its unprecedented scale in terms of images and object/action categories enables large-scale affordance training,
2) the diversity of scenarios and views ensures broad coverage of real-world interaction contexts,
3) precise annotation of affordable points eliminates the ambiguity of the mask/region boundary and aligns better with robotic execution. 
Additionally, we provide a benchmarking evaluation set for standardized comparisons in multi-modal affordance reasoning.

Based on HOVA-500K, we present \textbf{GLOVER++}, an end-to-end framework that explores distilling actionable affordance knowledge from raw human demonstrations for multiple robotic tasks as Fig.~\ref{fig:pipeline} shows. 
To achieve the balance between global semantic perception and local affordance learning, a \textit{global-to-local} affordance tuning policy is proposed to incorporate affordance reasoning capabilities while preserving the semantic understanding of VLM. 
We aim to unleash the potential of the affordance representation and
push the boundaries of affordance transfer in diverse manipulation tasks. Extensive experiments in both simulation and the real world are conducted to demonstrate the effectiveness of GLOVER++, including functional zero-shot manipulation, multi-task imitation learning, and serving as an effective perception module for long-horizon and bimanual manipulation.

Our contributions are summarized as follows:
\textbf{1}) We contribute a large-scale affordance-annotated dataset—HOVA-500K, that provides the necessary scale and diversity to learn generalizable affordance representations.
\textbf{2}) We present GLOVER++, a global-to-local paradigm of affordance training policy based on HOVA-500K, showing fine-grained affordance representation and generalizable affordance reasoning capability. GLOVER++ achieves state-of-the-art performance in the HOVA-500K evaluation benchmark. \textbf{3}) Extensive applications in tasks like zero-shot manipulation, multi-task imitation learning, long-horizon and bimanual manipulation demonstrate the huge potential of HOVA-500K and GLOVER++.

\begin{figure}[htbp]
  \centering
  \begin{minipage}[b]{0.38\textwidth}
    \centering
    \includegraphics[width=\linewidth]{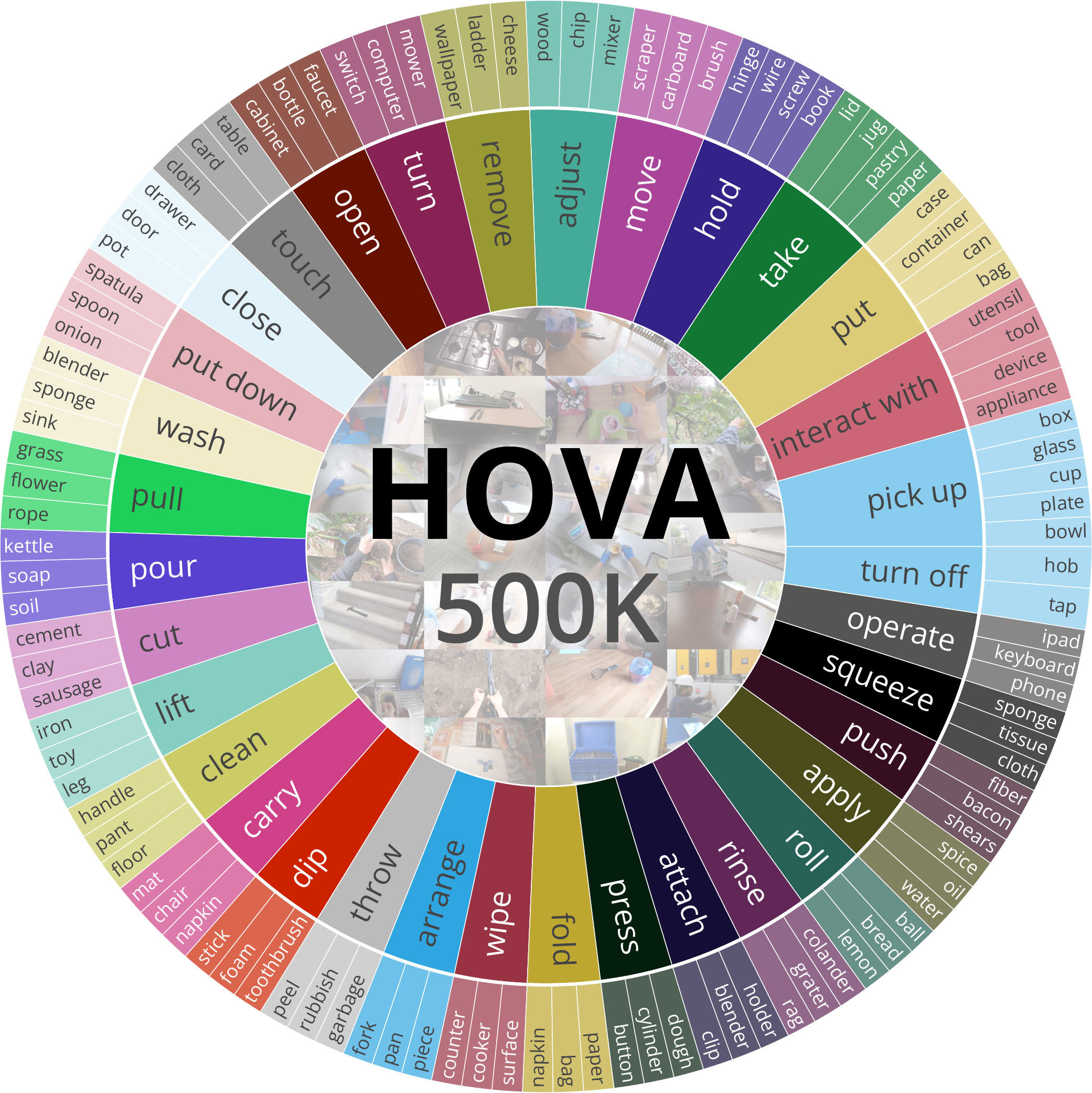} 
    \caption{The distribution of primary action categories ($>$1,000 samples) and related objects in HOVA-500K.}
     \label{fig:data}
  \end{minipage}
  \hfill
  \begin{minipage}[b]{0.6\textwidth}
    \centering
    \large
    \renewcommand{\arraystretch}{1}
    \resizebox{1.0\linewidth}{!}{
   \begin{tabular}{l|ccc|cccc}
      \hline
      \Xhline{1.\arrayrulewidth}
\textbf{Dataset} & \textbf{Img} & \textbf{Obj} & \textbf{Act} & \textbf{Format} & \textbf{Ego.} & \textbf{Exo.} & \textbf{Ann.} \\ \hline
 UMD \cite{myers2015affordance} & 30K & 17 & 7 & RGBD & \xmark & \cmark & Part  \\
 Sawatzky~\cite{sawatzky2017weakly} & 3K & 17 & 7 & RGBD & \xmark & \cmark & Part \\
 AGD20k \cite{agd20k} & 26K & 50 & 36 & RGB & \cmark & \cmark & Part   \\
 HANDAL \cite{guo2023handal} & 200K & 17 & - & RGBD & \xmark & \cmark & Obj/Part  \\
 OPRA \cite{fang2018demo2vec} & 11K & - & 7 & RGB & \cmark & \cmark & Point  \\
 AED \cite{li2024learning} & 0.7K & 13 & 8 & RGB & \xmark & \cmark & Part  \\
 3DOI \cite{3doi} & 10K & - & - & RGB & \cmark & \cmark & Point  \\
 PAD \cite{luo2021one} & 4K & 72 & 31 & RGB & \xmark & \cmark & Obj  \\
 IIT-AFF \cite{nguyen2017object} & 8.8K & 10 & 9 & RGBD & \xmark & \cmark & Part  \\ 
 ADE-Aff \cite{chuang2018learning} & 10K & 150 & 7 & RGB & \cmark & \xmark & Scene \\
\hline
\rowcolor{gray!25}
HOVA-500K & 500K & 1726 & 675 & RGB & \cmark & \cmark & Point  \\
\hline
\Xhline{1.\arrayrulewidth}
\end{tabular}
}
    \captionof{table}{Comparisons between HOVA-500K and previous datasets.  “\textbf{Format}”, “\textbf{Ann.}”,  “\textbf{Ego.}” and “\textbf{Exo.}” refer to the image format, egocentric, exocentric, and annotation type, respectively. Our HOVA-500K annotates the action \& object categories, and the affordance with more precise affordable points.} 
    \label{tab:example}
  
  \end{minipage}
 
\vspace{-10pt}
\end{figure}
\section{Related Works}

\noindent \textbf{Affordance Reasoning.}
Prior approaches to affordance inference can be categorized into three paradigms: (1) human-object interaction analysis~\cite{hassan2016attribute,hou2021affordance,luo2022learning}, (2) scene understanding through geometric and semantic cues~\cite{lerftogo,lu2024manigaussian,geff}, and (3) 3D point cloud grounding~\cite{geng2023rlafford,ning2024where2explore,wu2024learning,mo2021where2act}.
To achieve open-vocabulary affordance reasoning, recent advances have incorporated foundation models (LLMs/VLMs) to the model design~\cite{affcorrs, affordancellm, roboabc, song2023learning, lu2023vl, xu2023joint}. 
 This integration approach with LLMs/VLMs comprises two primary types. The first~\cite{roboabc, kuang2024ram} constructs a memory of object affordances and reasons about the affordances of objects in novel scenes by retrieving from the affordance memory with the help of CLIP~\cite{clip}.  On the other hand, methods like
 AffordanceLLM~\cite{affordancellm} and GLOVER~\cite{ma2024glover} fine-tune large VLMs~\cite{llava} on affordance datasets, leveraging the world knowledge and reasoning capability of foundation models. 
However, these approaches are constrained by limited data availability and insufficient exploration of the VLM-based affordance fine-tuning mechanism.
We contribute the HOVA-500K dataset to mitigate data scarcity, while providing an effective fine-tuning framework template, GLOVER++, to leverage such large-scale data.

\noindent \textbf{Language-guided Zero-shot Manipulation.}
Integrating linguistic modalities into robotic manipulation tasks serves as a crucial approach for achieving zero-shot manipulation capabilities~\cite{huang2023voxposer, huang2024rekep, di2024dinobot,okrobot, lerftogo, wang2024arm}.
Currently, the related methods can be categorized into multiple types like visual-language-action (VLA) pretraining in robotic data~\cite{brohan2023rt, rtx, kim2024openvla, team2024octo, driess2023palm}, invoking LLMs/VLMs for planning or in-context learning~\cite{ahn2022can, liang2023code, huang2023voxposer, huang2024rekep}, and using VLMs for object/scene representations~\cite{geff, lerf, lu2024manigaussian, lerftogo, shen2023distilled}. 
We follow the manner of leveraging VLMs for providing semantic scene understanding for the downstream manipulation tasks in this work. Different from the methods that distill features from 2D foundation models for building 3D feature fields~\cite{lerftogo, shen2023distilled} via neural rendering~\cite{nerf, gaussiansplat}, we adopt a visual-linguistic affordance representation and project the inferred affordance into 3D space without requiring full reconstruction.

\section{HOVA-500K Benchmark}

\noindent \textbf{Data Collection.}
HOVA-500K is primarily derived from three key sources: (1) human demonstration videos, which provide real-world interaction sequences for natural and diverse affordance learning (Ego4D~\cite{grauman2022ego4d}, EPIC-KITCHEN~\cite{epickitchen}); (2) object-part segmentation masks, offering structural mask annotations to bridge semantic parts with actionable regions (HANDAL~\cite{guo2023handal}); and (3) existing affordance datasets with labels of human-object affordable point (3DOI~\cite{3doi}). 
These datasets cover a broad range of scenarios, including both in-the-wild and household environments from ego/exo-centric views. 
This intentional diversity ensures robust generalization across different spatial relationships and interaction modalities.

\begin{wraptable}{r}{0.4\textwidth}
\vspace{-15pt}
  \centering
    \includegraphics[width=0.4\textwidth]{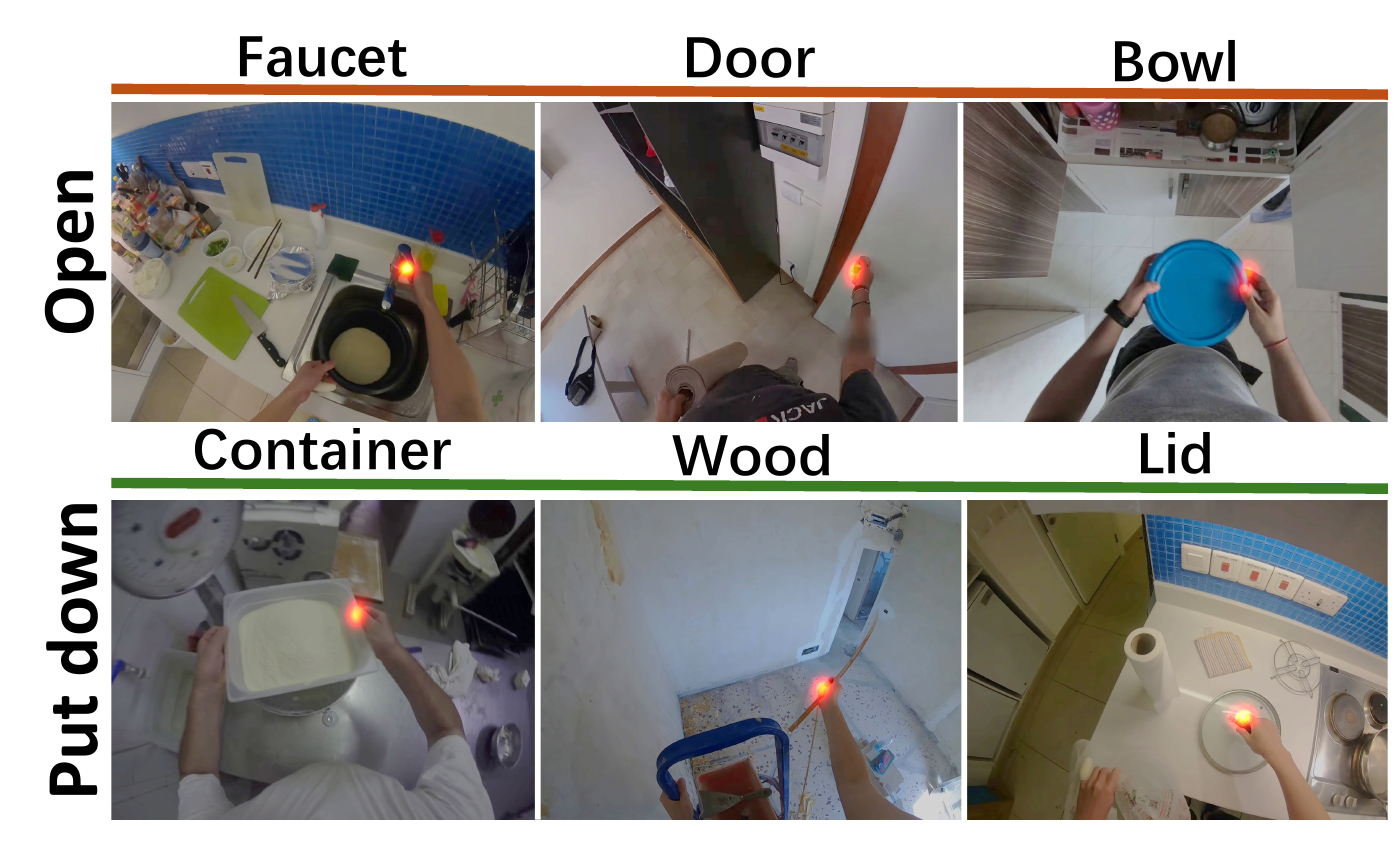}
   \captionsetup{type=figure}
   \vspace{-10pt}
   \caption{Some examples of HOVA-500K, showing action, object category, and Gaussian-distributed mask of affordable point.}
   \label{fig:data_exp}
    \vspace{-10pt}
\end{wraptable}

\noindent \textbf{Affordable Points Annotation.}
Unlike previous affordance datasets that annotate object regions as segmentation masks~\cite{guo2023handal}, we reformulates affordance learning as a dense keypoint prediction task, where the model predicts a single maximum-probability interaction point per object based on the functionality. This shift from region-level to point-level representation offers two advantages: (1) it eliminates the ambiguity of mask boundaries in precision-sensitive tasks, and (2) it better aligns with robotic control, where end-effector contact requires millimeter-level accuracy.

\noindent \textbf{Locating Affordable Points in Human Videos.}
Our approach builds upon the OCT model~\cite{oct}, applying skin segmentation~\cite{saxen2014skin} in the overlapping region between hand and object bounding boxes to obtain contact points. 
We then compute homography matrices between adjacent frames via RANSAC, enabling us to project contact points from the interaction frame back to the initial frame.
This step effectively eliminates occlusion caused by hands.
This semi-automatic pipeline enables efficient and scalable annotation of affordable points.
The specific introduction and pipeline is shown in Sec.~\ref{sec:points_collection} and Fig.~\ref{fig:data_pipeline}.

\noindent \textbf{Category \& Action.}
For the existing uni-modal visual affordance datasets, we implement a semi-automated labeling pipeline using Qwen-2.5-VL-7B~\cite{yang2024qwen2} VLM to generate object categories and actions. 
These automatically generated labels are subsequently verified by human annotators to eliminate clearly incorrect entries.
Representative actions and object categories are visualized in Fig.\ref{fig:data}, and more details can be found in Fig.~\ref{fig:data_vis_appendix}, \ref{fig:obj_ft_1000}, \ref{fig:action_ft_100}.

\noindent \textbf{Benchmarking Testing Set.}
\label{sec:testset}
To comprehensively evaluate affordance prediction models, we construct a diverse test set by selecting 6,000 images from HOVA-500K. 
Our evaluation framework measures both prediction accuracy and functional realism.
The metrics we used for evaluation include Kullback-Leibler Divergence ($KLD$), Similarity ($SIM$), and Normalized Scanpath Saliency ($NSS$). 
Moreover, we introduce $SIM_{part}$, a new metric that quantifies the practical plausibility of predicted affordance regions in real-world settings.
Detailed metric descriptions appear in Sec.~\ref{sec:Evaluation Metrics}.

\section{Method}
In this section, we elaborate on the training policy for distilling actionable affordance knowledge from human behaviors and aligning with human instructions. First, we briefly describe the task and preliminary (Sec.~\ref{sec:task_description}). Then, we introduce the global-to-local affordance fine-tuning of GLOVER++ in Sec.~\ref{sec:global-to-local}.
The potential application of GLOVER++ is elaborated in Sec.~\ref{sec:potential_applications}.

\subsection{Task Description and Preliminaries}
\label{sec:task_description}

GLOVER++ aims to predict executable affordable points $\mathcal{P}^{2D}$ in an open-vocabulary and end-to-end manner. Given the input $\mathcal{I}=(\mathcal{I}_R, \mathcal{T})$, where 
$\mathcal{I}_R, \mathcal{T}$ represent an RGB image and language instructions, we expect the model to generate $\mathcal{P}^{2D}$.
To achieve this, we first convert ground-truth affordable points $\mathcal{P}_{gt}^{2D}$ into Gaussian-distributed heatmaps $\mathcal{M}_{gt}^{2D}$ that centered at $\mathcal{P}_{gt}^{2D}$ to ensure gradient continuity during training like annotations in Fig.~\ref{fig:data_exp}. This translates discrete annotations into continuous optimization targets as shown in Fig.~\ref{fig:data_exp}. 
The resulting $\mathcal{M}_{gt}^{2D}$ supervises the model to generate affordance mask $\mathcal{M}_{\mathcal{A}}^{2D}$, where each pixel value $\mathcal{A}^{2D}$ in  $\mathcal{M}_{\mathcal{A}}^{2D}$ represents the affordable probability of the current position. 
We obtain $\mathcal{P}^{2D}$ by selecting the pixel with the highest probability: $\mathcal{P}^{2D} = \underset{\mathcal{P} \in \mathcal{I}_R}{\arg\max} \mathcal{M}_{\mathcal{A}}^{2D}$. This point is then projected into 3D space via camera intrinsics to yield $\mathcal{P}^{3D}$ for robotic execution.

\subsection{Global-to-Local Affordance Tuning}
\label{sec:global-to-local}

\begin{figure}[htbp]
\vspace{-10pt}
    \centering
    \includegraphics[width=1.0\linewidth]{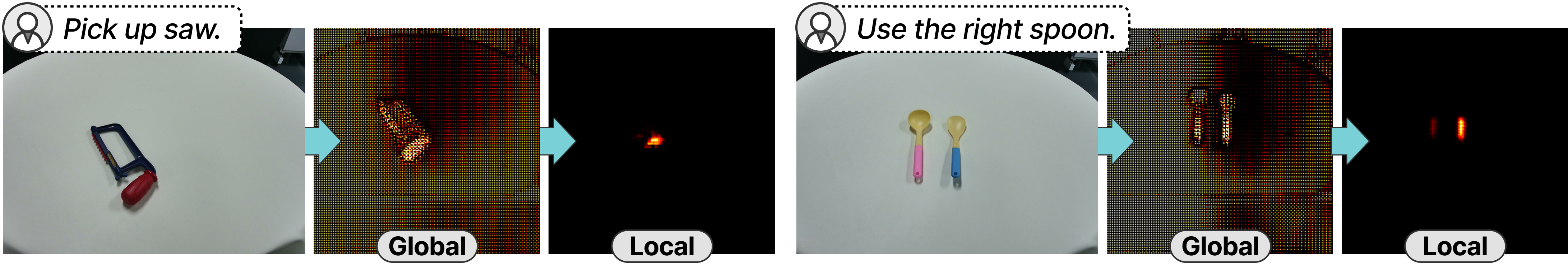}
    \caption{\textbf{Visualization of the decoded features by the global and local decoder} (the intensity of highlight scale with interest of regions). 
    We can observe that the integration of the local decoder effectively eliminates the background noise from the global decoding. 
    }
    \label{fig:global-to-local}
\end{figure}

We illustrate the global-to-local pipeline in Fig.~\ref{fig:pipeline} \& \ref{fig:glover++_pipeline}.
The following provides detailed descriptions of the components involved.

\noindent \textbf{Multi-modal Encoding.}
To empower GLOVER++ with world knowledge and reasoning capability, we leverage a VLM to encode the multi-modal inputs $\mathcal{I}$ into a hidden latent token for affordance reasoning. 
Following the \textit{Embedding-as-Mask} paradigm in LISA~\cite{lai2024lisa}, we add a new affordance token \texttt{<AFF>} to encode combined visual and linguistic features with LLaVA-1.5~\cite{llava}. Given the language instruction $\mathcal{T}$ and RGB image $\mathcal{I}_R$, we feed them into the LLaVA model $\mathbf{F}_{LLaVA}$ to generate the responsive hidden latents $\boldsymbol{r}$, from where we detach the latent \texttt{<AFF>} token:
    \begin{equation}
        \texttt{<AFF>} \in \boldsymbol{r} = \mathbf{F}_{LLaVA} (\mathcal{I}_R, \mathcal{T}).
    \end{equation}


\noindent \textbf{Global-to-Local Decoding.}
The \texttt{<AFF>} token aggregates \textbf{global} semantic context from $\mathcal{I_R}$, while the affordance prediction requires \textbf{local} fine-grained reasoning. 
Hence, the core challenge lies in \textit{balancing global semantic perception and precise local affordance representation learning}. 
To this end, we decompose the decoding process of \texttt{<AFF>} token into two stages: \textbf{global decoding} and \textbf{local decoding}.
In the first stage, the \texttt{<AFF>} token guides global semantic decoding to generate a high-level semantic logits map that captures global contextual relationships.
In the second stage, we refine the prediction through localized decoding: the semantic map $\mathcal{M}_{sem}^{2D}$ acts as a mask prompt to condition attention on relevant regions.
This enables accurate region-specific affordance prediction:
    \begin{equation}
            \mathcal{M}_{sem}^{2D} = \mathbf{F}_{dec}^{glo}(\texttt{<AFF>}, \boldsymbol{v}), \quad
        \mathcal{M}_{\mathcal{A}}^{2D} = \mathbf{F}_{dec}^{loc} (\mathcal{M}_{sem}^{2D}, \boldsymbol{v}),
    \end{equation}
where $\boldsymbol{v}$ denotes the visual features from the vision backbone. The effectiveness of the global-to-local decoding mechanism is visually demonstrated in Fig.~\ref{fig:global-to-local}, and implementation details are specified in Sec.~\ref{sec:model_appendix}.
 
\noindent \textbf{Training Objective.}
Besides the sigmoid focal loss~\cite{lin2017focal} used in GLOVER~\cite{ma2024glover}, 
we introduce an additional Kullback-Leibler Divergence (KLD) loss to constrain the predicted affordance distribution.
\begin{wraptable}{r}{0.3\textwidth}
\vspace{-5pt}
  \centering
    \includegraphics[width=0.3\textwidth]{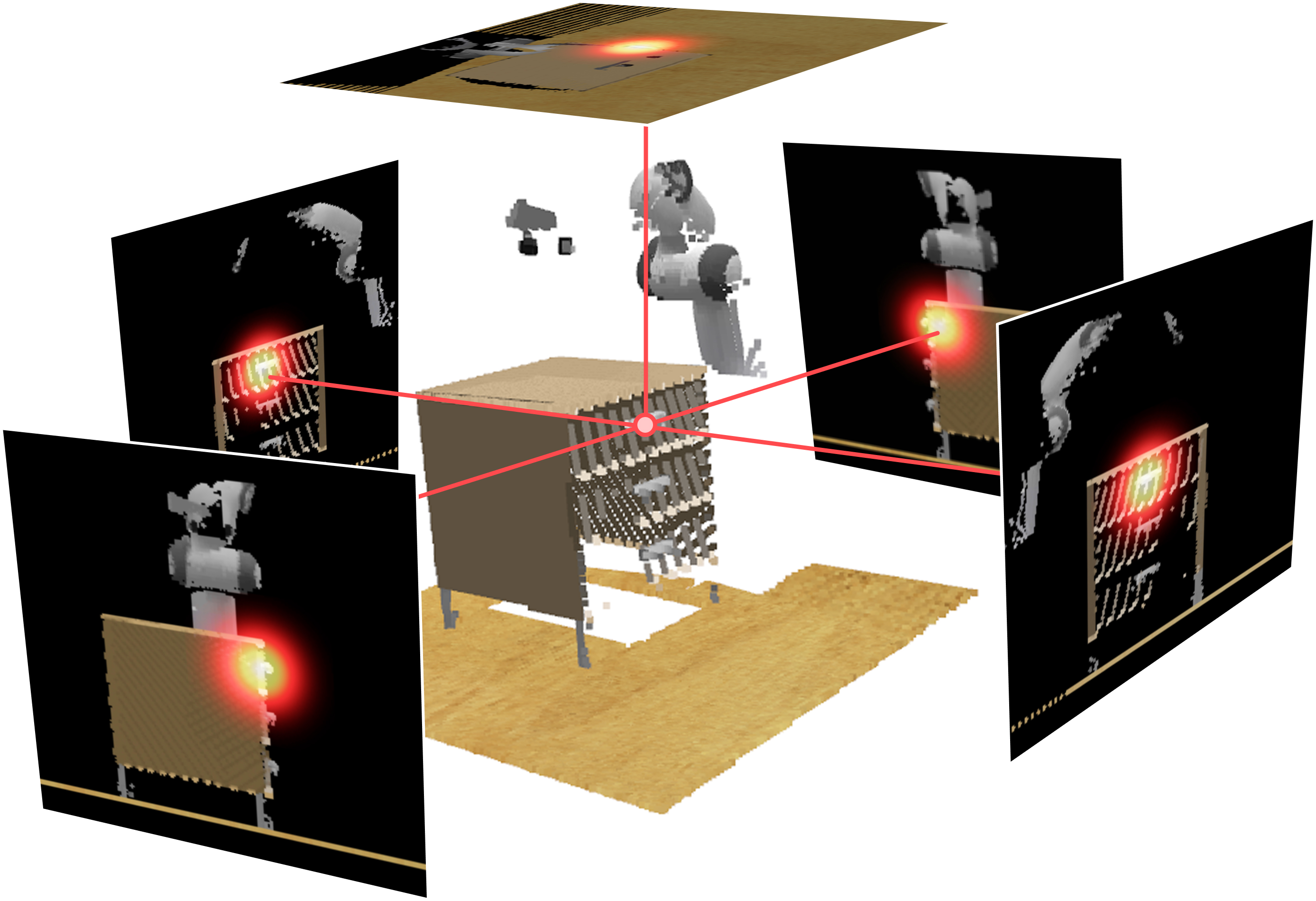}
   \vspace{-10pt}
   \captionsetup{type=figure}
   \caption{Explicit affordance representation for imitation learning in RLBench~\cite{james2020rlbench}.}
   \label{fig:rlbench_vis}
    \vspace{-20pt}
\end{wraptable}
 This KLD term aligns the predicted heatmap $\mathcal{M}_{\mathcal{A}}^{2D}$ with the Gaussian-distributed ground truth $\mathcal{M}_{gt}^{2D}$, encouraging distributional consistency.
 The overall training objective becomes:
\begin{equation}
    \mathcal{L}=\mathcal{L}_{FL}(\mathcal{M}_{\mathcal{A}}^{2D}, \mathcal{M}_{gt}^{2D})+\mathcal{L}_{KL}(\mathcal{M}_{\mathcal{A}}^{2D}, \mathcal{M}_{gt}^{2D}),
\end{equation}
More details about the training objective are specified in Sec.~\ref{sec:Training Objective}.

\subsection{Unleash the Potential of Affordance Representation}
\label{sec:potential_applications}

\noindent \textbf{Zero-shot Manipulation.}
GLOVER++ is capable of reasoning about affordance in an open-vocabulary way to acquire 3D graspable points, 
which inherently addresses zero-shot grasping challenges.
The inferred affordable points $\mathcal{A}^{3D}$ can be combined with all kinds of pose estimators (\textit{e.g.} GraspNet~\cite{graspnet}, AnyGrasp~\cite{anygrasp}, FoundationPose~\cite{wen2024foundationpose}) or geometric-constraints estimation (\textit{e.g.} superquadric recovery~\cite{leonardis1997superquadrics, paschalidou2019superquadrics, ma2024glover}) to generate the grasping pose $(\mathcal{A}^{3D}, \mathcal{\tau}^{3D})$, where $\mathcal{\tau}^{3D}$ is the rotation of the end-effector. For motion planning, we use Inverse Kinematics (IK) by default to reach $\mathcal{A}^{3D}$.

\noindent \textbf{Imitation Learning.}
Instead of the previous \textit{pretraining-finetuning} paradigm that transfers pretrained weights in human videos to downstream imitation-learning tasks~\cite{nair2022r3m, xiao2022mvp, zhou2024mitigating}, we adopt the reasoned affordance representation as an explicit knowledge prior to dynamically modulate attention weights, enabling the model to focus on task-relevant regions, as shown in Fig.~\ref{fig:rlbench_vis}. 
This representation serves as a structured guidance signal, making the learning process more interpretable and effective compared to implicit methods. 
The pipeline is shown in Fig.~\ref{fig:rlbench_pipeline}, and training details are specified in Sec.~\ref{sec:IL_appendix}. 



\section{Experiments}






We evaluate the performance of GLOVER++ from four perspectives: vision-language affordance reasoning (Sec.~\ref{sec:vl_affordance}), zero-shot manipulation (Sec.~\ref{sec:zero-shot-mani}), imitation learning (Sec.~\ref{sec:il-exp}), and extended capabilities (Sec.~\ref{sec:exp-other}), including long-horizon manipulation with VLM planner and bimanual manipulation.

\vspace{-5pt}
\begin{figure}[h]
    \centering
    \includegraphics[width=1.0\textwidth]{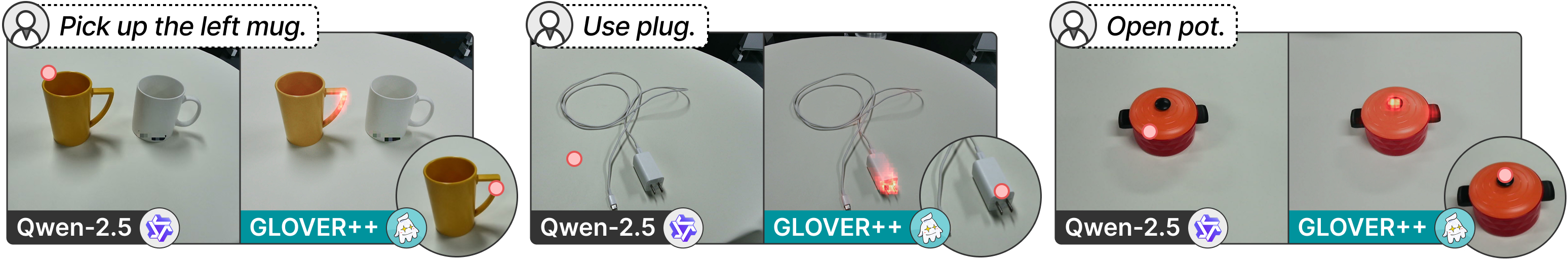}
    \caption{\textbf{Comparison with Qwen-2.5-VL}, 
    GLOVER++ generates more physically plausible and functionally grounded prediction results, aligning better with real-world interaction constraints.}
    \vspace{-5pt}
    \label{fig:compare-to-qwen}
\end{figure}

\subsection{Vision-Language Affordance Reasoning}
\label{sec:vl_affordance}


\noindent \textbf{Training Details.} 
GLOVER++ is trained on 8 NVIDIA A6000 GPUs for 10 epochs with a batch size of 32 per GPU. 
We employ AdamW~\cite{loshchilov2017decoupled} optimizer with a weight decay of 5e-4. The learning rates for the global decoder and local decoder are set to 5e-5 and 5e-4 to achieve a balance between preserving open-vocabulary knowledge and affordance learning.   For more details, please refer to Sec.~\ref{sec:training_details}.


\begin{wraptable}{r}{0.55\textwidth}
\vspace{-10pt}
  \centering
  \renewcommand{\arraystretch}{1.2}
  \resizebox{0.55\textwidth}{!}{
   \begin{tabular}{ccccc}
      \toprule
      \textbf{Methods} & \textbf{KLD} $\downarrow$ & \textbf{SIM} $\uparrow$ & $\textbf{SIM}_{part}$ $\uparrow$ & \textbf{NSS} $\uparrow$ \\
      \hline
      3DOI~\cite{3doi} & 5.978 & 0.007 & 0.006 & -0.311 \\
      AffordanceLLM~\cite{affordancellm}  & 5.041 & 0.018  & 0.161 & 1.665 \\
    GLOVER~\cite{ma2024glover}  & 4.874 & 0.016 & 0.254 & 2.876 \\
      \rowcolor{gray!20}
      \textbf{GLOVER++} & \textbf{3.411} & \textbf{0.141} & \textbf{0.563} & \textbf{5.296} \\ \hline

     & \multicolumn{4}{c}{\textbf{\textit{Ablations}}} \\\cmidrule(r){2-5}
      w/o global-to-local & 3.465 & 0.101 & 0.483 & 4.925 \\
      w/o KLD & 4.307 & 0.030 & 0.409 & 4.005 \\
      10$\rightarrow$5 epoch & 3.615 & 0.121 & 0.533 & 5.197 \\
      
      \bottomrule
   \end{tabular}
   }
   \caption{Affordance reasoning results and ablations in the benchmarking dataset. 
   }
   \label{tab:affordance}
  \vspace{-10pt}
\end{wraptable}

\noindent \textbf{Results.}
Table~\ref{tab:affordance} shows the quantitative results of affordance reasoning on the HOVA-500K benchmark. The metrics specified in the benchmarking testing set (Sec.~\ref{sec:testset}) are adopted to evaluate.
We compare GLOVER++ with three methods, 3DOI~\cite{3doi}, AffordanceLLM~\cite{affordancellm}, and GLOVER~\cite{ma2024glover}.
The three methods are all based on affordance pretraining. Specifically, 3DOI relies solely on visual inputs. In contrast, both AffordanceLLM and GLVOER are pretrained models that integrate visual-language information for affordance learning.
GLOVER++ significantly outperforms all baselines across all metrics, owing to its global-to-local decoding scheme and KLD-based optimization, which jointly enhance both affordance center prediction and distributional alignment.

We also benchmark against Qwen-2.5-VL-7B~\cite{yang2024qwen2}, a powerful VLM with strong spatial understanding via Rotary Positional Embedding (RoPE).
As shown in Fig.~\ref{fig:compare-to-qwen}, while Qwen-2.5-VL exhibits decent localization, GLOVER++ provides affordance predictions that are more physically plausible and functionally grounded, despite using a comparable model size ($\sim$7B).

\noindent \textbf{Ablations.} 
To assess the contribution of key components, we evaluate ablations by removing the global-to-local module, the KLD loss, and varying the training length (Table~\ref{tab:affordance}).
Notably, removing global-to-local decoding or KLD optimization results in a $28.4\%$ and $78.7\%$ drop in the SIM metric, respectively, underscoring their importance. We also show the visualization of reasoned affordance in Fig.~\ref{fig:ablation_compare_vis}, where the effectiveness of model components is evident. Additional ablations are provided in Sec.~\ref{sec:ablation_appendix}.

\subsection{Zero-shot Manipulation}
\label{sec:zero-shot-mani}
\noindent \textbf{Setup.} We perform extensive experiments in both the simulated and real-world settings. For simulation,  we use IsaacGym~\cite{makoviychuk2021isaac} with 
GAPartNet~\cite{geng2023gapartnet} object sets, 
\begin{wraptable}{r}{0.55\textwidth}
\vspace{-5pt}
  \centering
  \resizebox{0.55\textwidth}{!}{
  \begin{tabular}{ccccccccccc}
    \toprule
    Object 
    & \multicolumn{2}{c}{\includegraphics[width=0.06\linewidth]{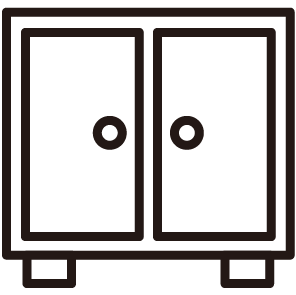}} 
    & \multicolumn{2}{c}{\includegraphics[width=0.06\linewidth]{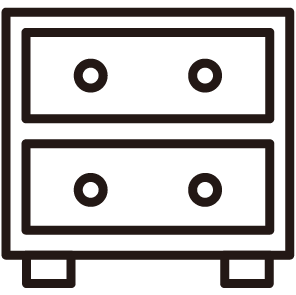}} 
    & \multicolumn{1}{c}{\includegraphics[width=0.06\linewidth]{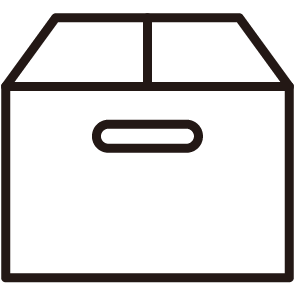}} 
    & \multicolumn{1}{c}{\includegraphics[width=0.06\linewidth]{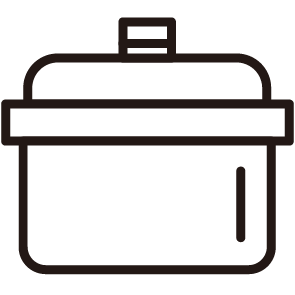}} 
    & \multicolumn{1}{c}{\includegraphics[width=0.06\linewidth]{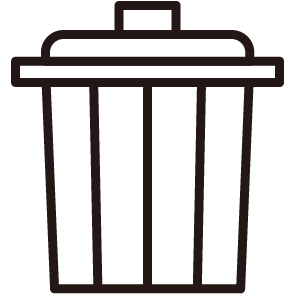}} 

    & \textbf{AVG} \\
    \cmidrule(r){1-9}
    Task & \texttt{O} & \texttt{C} & \texttt{O} & \texttt{C} & \texttt{O} & \texttt{P} & \texttt{O} &/ \\
    \midrule
    VRB~\cite{vrb} &4 &60 &4 &56 &20 &24 & 16 &26.3\\
    Robo-ABC~\cite{roboabc} &28  &44  &32 &32  &20   &32 &16 &29.1 \\
    RAM~\cite{kuang2024ram} &36 &\textbf{64} &40 &60 & 24 & 36 &32 &41.7 \\
    \textbf{GLOVER++} &\textbf{40} &60 &\textbf{40} &\textbf{68} &\textbf{32} &\textbf{44} &\textbf{44} &\textbf{46.9}\\
    \bottomrule
  \end{tabular}
  }
  \caption{Success rates of different methods in IsaacGym. \texttt{O}, \texttt{C}, \texttt{P} represent \textit{Open, Close, Pickup}, respectively.}
  \label{tab:zero-shot-sim}
  \vspace{-15pt}
\end{wraptable}
involving a 7-DoF Franka Panda arm for zero-shot manipulation.
50 objects across 5 categories (\textit{Box, Pot, Drawer, TrashCan, Cabinet}) are used. 
Actions are listed in Table~\ref{tab:zero-shot-sim}, with success defined as articulation joint or height exceeding a predefined threshold. Each task is tested 25 times with varied initial poses.
For real-world experiments, we deploy a 7-DoF UFactory xArm, evaluated over five trials per task using RGB-D input from an Orbbec Femto Bolt ($1280 \times 960$ resolution). Object categories and actions are provided in Table~\ref{tab:zero-shot-real}. Note that the ``\textit{Press Button}" task involves discriminating the correct color one and pressing it.

\begin{wraptable}{r}{0.5\textwidth}
\vspace{-15pt}
  \centering
  \resizebox{0.5\textwidth}{!}{
  \begin{tabular}{cccccccccc}
    \toprule
    Object 
    & \multicolumn{1}{c}{\includegraphics[width=0.06\linewidth]{secs/icon/Drawer.png}} 
    & \multicolumn{1}{c}{\includegraphics[width=0.06\linewidth]{secs/icon/Pot.png}} 
    & \multicolumn{1}{c}{\includegraphics[width=0.06\linewidth]{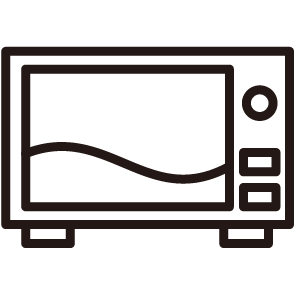}} 
    & \multicolumn{1}{c}{\includegraphics[width=0.06\linewidth]{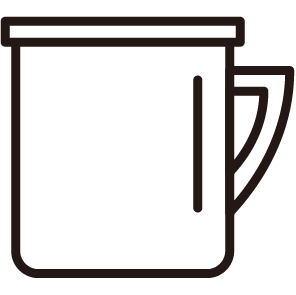}} 
    & \multicolumn{1}{c}{\includegraphics[width=0.06\linewidth]{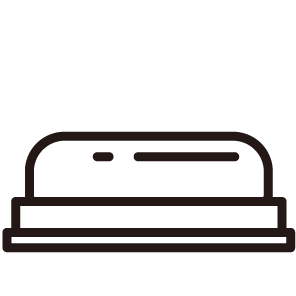}} 
    & \multicolumn{1}{c}{\includegraphics[width=0.06\linewidth]{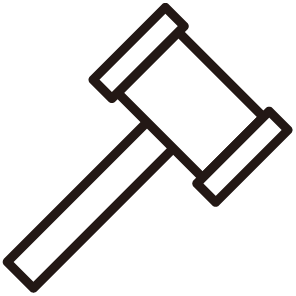}} 
    & \textbf{AVG} \\
    \cmidrule(r){1-8}
    Task & \texttt{O} & \texttt{P} & \texttt{O} & \texttt{P} & \texttt{PR} & \texttt{P} & / \\
    \midrule
    RAM~\cite{kuang2024ram} & 3/5 & 1/5 & 2/5 & 3/5 & 1/5 & 4/5 & 46.7  \\
    
    \textbf{GLOVER++} & \textbf{4/5} & \textbf{4/5} & 2/5 & \textbf{4/5} & \textbf{4/5} & 4/5 & \textbf{73.3} \\
    \bottomrule
  \end{tabular}
  }
  \caption{Success rates of different methods in real-world experiments. \texttt{O, P, PR} means \textit{Open, Pickup, Press}, respectively.}
  \label{tab:zero-shot-real}
  \vspace{-10pt}
\end{wraptable}
\noindent \textbf{Baselines \& Results.} We compare GLOVER++ with three baselines. 
VRB~\cite{vrb} predicts contact points by learning from human demonstrations. Both Robo-ABC~\cite{roboabc} and RAM~\cite{kuang2024ram} retrieve affordance from the pre-built memory and transfer it to new scenes via estimated similarity of CLIP~\cite{clip}. 
For simulation and real-world experiments, we use the success rate (SR) as metric. 
Table~\ref{tab:zero-shot-sim} and \ref{tab:zero-shot-real} show the simulated and real-world results, respectively. We can see that GLOVER++ achieves favorable performance among the baseline methods in both the IsaacGym and real-world environments, yielding an average success of $46.9\%$ and $73.3\%$, respectively. 
Compared to GLOVER++, RAM is limited in its ability to distinguish object properties (such as buttons of different colors) due to the retrieval mechanism based on object-level similarity. 
In contrast, GLOVER++ generalizes to novel objects and scenarios by directly grounding actionable affordance points from language instructions alone.

\subsection{Imitation Learning}
\label{sec:il-exp}
\noindent \textbf{Setup.} 
We validate affordance knowledge transfer in language-guided multi-task imitation learning using RLBench~\cite{james2020rlbench}.
RLBench is a robot manipulation benchmark built on CoppelaSim~\cite{rohmer2013v} and PyRep~\cite{james2019pyrep}. 
We follow the protocols of PerAct~\cite{peract2022arxiv} to test the model on 18 tasks in RLBench by controlling a Franka Panda robot with a parallel gripper. 
Each policy is trained on 100 demonstrations using RGB-D observations from four views (front, left shoulder, right shoulder, wrist) at $128 \times 128$ resolution. Tasks are tested 25 times per trial. Additional details are in Sec.~\ref{sec:IL_appendix}.

\begin{table*}[!th]
\vspace{-15pt}
\centering
\resizebox{0.95\linewidth}{!}{
\begin{tabular}{cccccccccccccccccccccc}
\specialrule{0.9pt}{0pt}{0pt}
\multicolumn{1}{c|}{\makecell[c]{Models}} 
                            &\begin{tabular}[c]{@{}c@{}}\texttt{put in} \\\texttt{drawer}\end{tabular}    
                            &\begin{tabular}[c]{@{}c@{}}\texttt{drag} \\\texttt{stick}\end{tabular} 
                            &\begin{tabular}[c]{@{}c@{}}\texttt{turn} \\\texttt{tap}\end{tabular}   
                            &\begin{tabular}[c]{@{}c@{}}\texttt{slide} \\\texttt{block}\end{tabular} 
                            &\begin{tabular}[c]{@{}c@{}}\texttt{open} \\\texttt{drawer}\end{tabular}    
                            &\begin{tabular}[c]{@{}c@{}}\texttt{put in}\\\texttt{cupboard}\end{tabular} 
                            &\begin{tabular}[c]{@{}c@{}}\texttt{sort}\\\texttt{shape}\end{tabular} 
                            &\begin{tabular}[c]{@{}c@{}}\texttt{put in}\\\texttt{safe}\end{tabular} 
                            &\begin{tabular}[c]{@{}c@{}}\texttt{push}\\\texttt{buttons}\end{tabular} 
                            &\begin{tabular}[c]{@{}c@{}}\texttt{close}\\\texttt{jar}\end{tabular}
                            \\
\specialrule{0.9pt}{0pt}{0pt}

\addlinespace[0.5ex]
\specialrule{0.5pt}{0pt}{0pt}
\multicolumn{1}{c|}{RVT} &92.0 &100.0 &\textbf{100.0} &76.0 &76.0 &52.0 &40.0 & 84.0 &96.0 &92.0\\
\multicolumn{1}{c|}{RVT-AFF} &92.0 &100.0 &96.0 &64.0 &76.0 &\textbf{76.0} &40.0 &\textbf{88.0} &96.0 &\textbf{96.0} \\
\specialrule{0.5pt}{0pt}{0pt}

\addlinespace[0.5ex]
\specialrule{0.5pt}{0pt}{0pt}
\multicolumn{1}{c|}{RVT2} & \textbf{96.0} &100.0 &100.0 &88.0 &68.0 &68.0 &32.0 &96.0 &100.0  &100.0 \\
\multicolumn{1}{c|}{RVT2-AFF} & 92.0 &100.0 &100.0 &\textbf{92.0} &\textbf{72.0} &68.0 &\textbf{36.0} &\textbf{100.0} &100.0 &100.0 \\
\specialrule{0.5pt}{0pt}{0pt}

\\[1pt]  
\specialrule{0.9pt}{0pt}{0pt}
\multicolumn{1}{c|}{\makecell[c]{Models}} 
                            &\begin{tabular}[c]{@{}c@{}}\texttt{stack} \\\texttt{blocks}\end{tabular}
                            &\begin{tabular}[c]{@{}c@{}}\texttt{place}\\\texttt{wine}\end{tabular} 
                            &\begin{tabular}[c]{@{}c@{}}\texttt{sweep to} \\\texttt{dustpan}\end{tabular}    
                            &\begin{tabular}[c]{@{}c@{}}\texttt{meat off} \\\texttt{grill}\end{tabular} 
                            &\begin{tabular}[c]{@{}c@{}}\cellcolor{successColor}{\texttt{screw}}\\\cellcolor{successColor}{\texttt{bulb}}\end{tabular} 
                             &\begin{tabular}[c]{@{}c@{}}\cellcolor{successColor}{\texttt{place}}\\\cellcolor{successColor}{\texttt{cups}}\end{tabular}
                            &\begin{tabular}[c]{@{}c@{}}\cellcolor{successColor}{\texttt{insert}}\\\cellcolor{successColor}{\texttt{peg}}\end{tabular}
                            &\begin{tabular}[c]{@{}c@{}}\cellcolor{successColor}{\texttt{stack}}\\\cellcolor{successColor}{\texttt{cups}}\end{tabular} 
                            
                            &\multicolumn{2}{p{2.5cm}}{\cellcolor{gray!20} \makecell[c]{Averaged \\ Success Rate}} 
                            \\
\specialrule{0.9pt}{0pt}{0pt}

\addlinespace[0.5ex]
\specialrule{0.5pt}{0pt}{0pt}
\multicolumn{1}{c|}{RVT} &24.0   &88.0    &64.0   &88.0 &\cellcolor{successColor}{48.0} &\cellcolor{successColor}{0} &\cellcolor{successColor}{0} &\cellcolor{successColor}{0} & \multicolumn{2}{p{2.5cm}}{\cellcolor{gray!20}\makecell[c]{$62.2$}} \\
\multicolumn{1}{c|}{RVT-AFF} & \textbf{32.0}   &\textbf{92.0}   &\textbf{88.0}    &\textbf{96.0} &\cellcolor{successColor}{\textbf{56.0}} &\cellcolor{successColor}{\textbf{4.0}} &\cellcolor{successColor}{\textbf{8.0}}  &\cellcolor{successColor}{\textbf{8.0}} & \multicolumn{2}{p{2.5cm}}{\cellcolor{gray!20}\makecell[c]{\textbf{$67.1 \myPink{(+4.9)}$}}} \\

\specialrule{0.5pt}{0pt}{0pt}

\addlinespace[0.5ex]
\specialrule{0.5pt}{0pt}{0pt}
\multicolumn{1}{c|}{RVT2} &\textbf{80.0}    & 92.0  &100.0   &96.0 & \cellcolor{successColor}{88.0} &\cellcolor{successColor}{36.0}  &\cellcolor{successColor}{40.0}  & \cellcolor{successColor}{72.0} & \multicolumn{2}{p{2.5cm}}
{\cellcolor{gray!20}\makecell[c]{$80.7$}}\\
\multicolumn{1}{c|}{RVT2-AFF}  &76.0   &\textbf{100.0}    &100.0    &96.0 &\cellcolor{successColor}{\textbf{92.0}}  &\cellcolor{successColor}{\textbf{44.0}} &\cellcolor{successColor}{\textbf{52.0}} &\cellcolor{successColor}{\textbf{80.0}} & \multicolumn{2}{p{2.5cm}}{\cellcolor{gray!20}\makecell[c]{\textbf{$83.3 \myPink{(+2.6)}$}}}\\
\specialrule{0.5pt}{0pt}{0pt}
\end{tabular}
}
\setlength{\abovecaptionskip}{0.2cm}
\caption{\textbf{Success rate of 18 tasks in RLBench}. With explicit affordance representation, both RVT~\cite{goyal2023rvt} and RVT-2~\cite{goyal2024rvt2} show improvements in the multiple imitation-learning tasks.}
\vspace{-15pt}
\label{tab:rlbench}
\end{table*}

\noindent \textbf{Baselines \& Results.}
We compare with two imitation-learning baselines.  RVT~\cite{goyal2023rvt} utilizes a multi-view Transformer model to extract visual features from multi-view images rendered based on point clouds, and predicting end-effector poses via deep-learning-based Q-function estimation. RVT-2~\cite{goyal2024rvt2} learns more precise manipulations via zooming into the region of interest. 

As shown in Table~\ref{tab:rlbench}, GLOVER++ improves RVT and RVT-2 by \myPink{+4.9\%} and \myPink{+2.6\%}, respectively. Gains are particularly evident in tasks requiring precise control (\textit{e.g.}, \texttt{insert peg, stack cups}, \textcolor{blue}{blue} highlight in Table~\ref{tab:rlbench}). 
The affordance representation constrains attention to task-relevant spatial and semantic features, improving policy effectiveness.
The experiments demonstrate that the explicit representation like affordance is capable of transferring knowledge learned from human demonstrations to enhance robotic imitation learning performance.

\subsection{Extended Capabilities}
\label{sec:exp-other}

\begin{figure}[htbp]
\vspace{-5pt}
    \centering
    \includegraphics[width=1.0\linewidth]{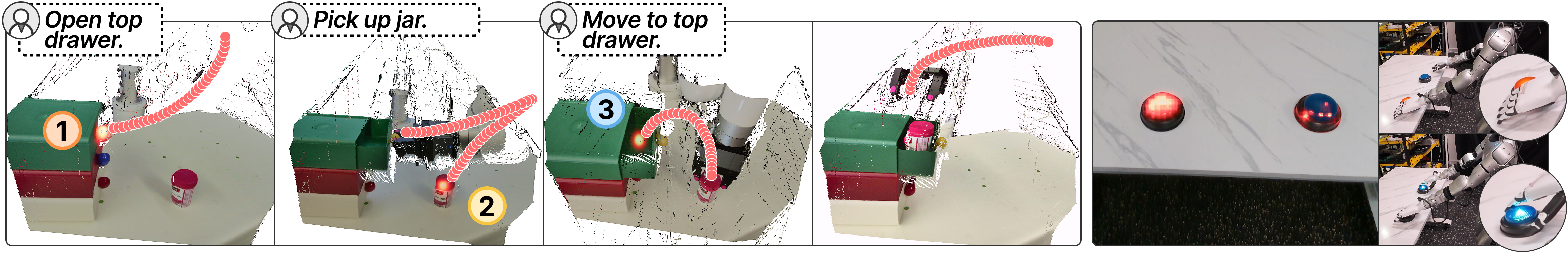}
    \caption{\textbf{Left:} GLOVER++ serves as a perceptual module for the VLM planner to complete long-horizon tasks. \textbf{Right:} GLOVER++ enables bimanual tasks by reasoning affordances for both left and right hands with spatial relationships.}
    \label{fig:extend}
\vspace{-5pt}
\end{figure}


\noindent \textbf{Long-horizon Manipulation with VLM planner.}
GLOVER++ can serve as a perceptual backbone for a high-level VLM planner. We integrate it with Qwen-2.5-VL~\cite{yang2024qwen2}, which decomposes long-horizon instructions into subgoals. 
As shown in Fig.\ref{fig:extend}-left, Qwen-2.5-VL split the task ``\textit{Put the jar into the top drawer}" into steps like ``\textit{Open top drawer}", ``\textit{Pick up jar}", ``\textit{Move to top drawer}" \textit{etc.}, and invoking GLOVER++ when affordance grounding is required (\ding{172}, \ding{173}, \ding{174}).
This hybrid system combines semantic planning and precise affordance prediction, enabling robust multi-stage manipulation. Full flow is shown in Fig.\ref{fig:qwen_planner}.

\noindent \textbf{Bimanual Manipulation.}
Thanks to its spatial reasoning capabilities, GLOVER++ can interpret positional cues (\textit{e.g.}, “left/right”, “top/bottom”) to enable dual-arm affordance reasoning. It generates graspable regions for both arms while maintaining spatial separation and feasibility (Fig.\ref{fig:extend}, right). We execute dual-arm motions using obstacle-avoidance IK on the Unitree G1 humanoid robot (Fig.\ref{fig:humanoid_avoid}).



\section{Conclusion}
\label{sec:conclusion}

In this work, we address the critical challenges of transferring actionable affordance knowledge from human demonstrations to robotic tasks by introducing HOVA-500K, a large-scale dataset with precise affordance annotations, and GLOVER++, a global-to-local framework for affordance reasoning. 
We demonstrate the potential of affordance representation and GLOVER++ in diverse tasks, including zero-shot manipulation, imitation learning, long-horizon and bimanual manipulation.
 We aim to foster future research in the explicit, interpretable and transferable representation learning for robotic manipulation from human behaviors.
 
\section{Limitations}
While HOVA-500K provides a large-scale affordance dataset, its annotations are primarily derived from static images, limiting coverage of dynamic interactions (e.g., tool-use trajectories or force-sensitive affordances). 
GLOVER++’s reliance on vision-language models may inherit biases from pre-trained VLMs, occasionally leading to over-generalized affordance predictions for novel object-action combinations. 
Additionally, GLOVER++ relies on imitation learning or an extra VLM planner to complete long-horizon manipulation tasks, lacking the ability to plan the grasping pose and trajectories by itself.
Future work to address these issues includes: enlarging the HOVA-500K with annotated trajectories of human behaviors, empowering GLOVER++ with trajectory planning ability via finetuning, and training multiple versions of GLOVER++ based on different large VLMs. We also discuss the \textbf{failure cases} in Sec.~\ref{sec:failure_case}.




\bibliography{example}  
\appendix


\section{HOVA-500K}

\subsection{Locate Affordable Points in Human Videos.}
\label{sec:points_collection}
\begin{figure}[htbp]
    \centering
    \includegraphics[width=0.9\linewidth]{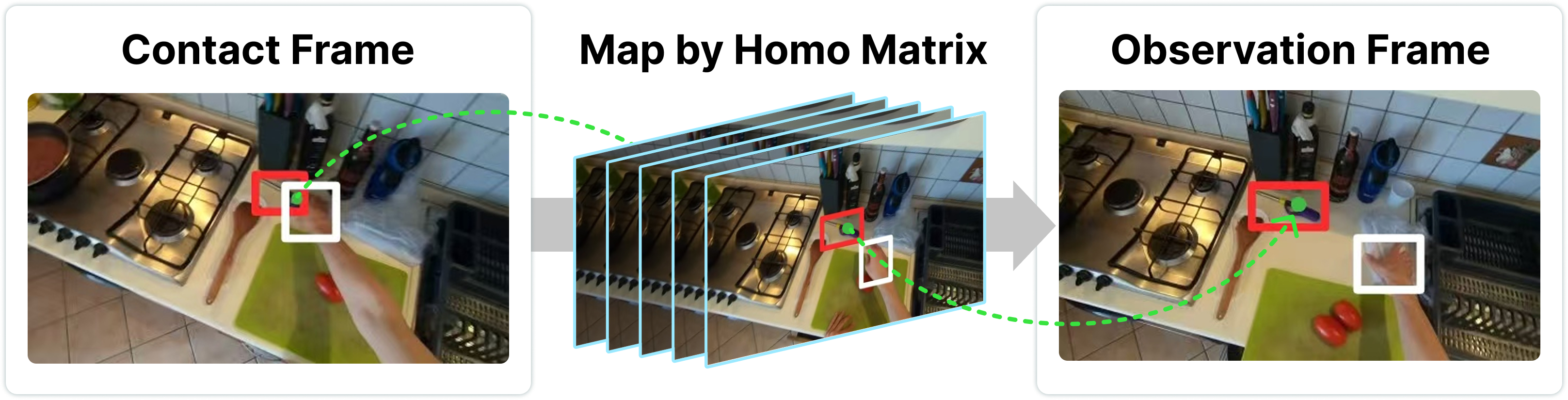}
    \caption{The pipeline of locating affordable points from the human videos following OCT~\cite{oct}. We initially apply skin segmentation~\cite{saxen2014skin} in the overlapping region between hand and object bounding boxes to obtain contact points, and compute the homography between each pair of successive frames. We project the contact points back to the initial frame based on the homography.}
    \label{fig:data_pipeline}
\end{figure}

Our approach builds upon the OCT model~\cite{oct}. For a given contact frame~$\mathcal{C}$, we initially apply skin segmentation~\cite{saxen2014skin} in the overlapping region between hand and object bounding boxes to obtain contact points $P=\{p_1,...,p_2\}$. 
Subsequently, we sample 10 preceding frames as observation key frames $\mathcal{O}=\{o_1,...,o_{10}\}$ (where $o_{10}$ is temporally adjacent to $\mathcal{C}$). 
We aim to calculate a set of transformation matrices that link all observation frames to $\mathcal{C}$ by computing the homography between consecutive frames.
To estimate those homographies, we mask out dynamic elements like detected hands and objects from each frame. 
Then, we establish feature correspondences in the unmasking regions using SURF descriptor~\cite{bay2006surf}. 
The homography is computed by applying RANSAC algorithm to the established feature correspondences. Finally, we project the contact points back to the initial frame $o_1$ based on estimated tomography to acquire affordable points.
The pipeline is shown in Fig.~\ref{fig:data_pipeline}.

\subsection{Benchmark Evaluation Metrics}
\label{sec:Evaluation Metrics}
We utilize four metrics to evaluate the models' performance in the HOVA-500K benchmark, including $KLD, SIM, NSS$, and $SIM_{part}$.
Specifically, these metrics can be formulated as:
\begin{equation}
    KLD(\mathcal{M}_{\mathcal{A}}^{2D},\mathcal{M}_{gt}^{2D})=\sum_{i}{{\mathcal{M}_{gt}^{2D}}_i\log{(\epsilon+\frac{{\mathcal{M}_{gt}^{2D}}_i}{\epsilon+{\mathcal{M}_{\mathcal{A}}^{2D}}_i})}}
\end{equation}
\begin{equation}
    SIM(\mathcal{M}_{\mathcal{A}}^{2D},\mathcal{M}_{gt}^{2D})=\sum_{i}{\min({\mathcal{M}_{gt}^{2D}}_i,{\mathcal{M}_{\mathcal{A}}^{2D}}_i)}
\end{equation}
\begin{equation}
    NSS(\mathcal{M}_{\mathcal{A}}^{2D},\mathcal{M}_{gt}^{2D})=\frac{1}{N}\sum_{i}{\mathcal{\hat{M}}_{\mathcal{A}}^{2D}\times\mathcal{M}_{gt}^{2D}}
\end{equation}
where $N=\sum_{i}{{\mathcal{M}_{gt}^{2D}}_i}$, $\hat{M}_{\mathcal{A}}^{2D}=\frac{\mathcal{M}_{\mathcal{A}}^{2D}-\mu(\mathcal{M}_{\mathcal{A}}^{2D})}{\sigma(\mathcal{M}_{\mathcal{A}}^{2D})}$, $\mu(\mathcal{M}_{\mathcal{A}}^{2D})$ and $\sigma(\mathcal{M}_{\mathcal{A}}^{2D})$ are the mean and standard deviation, respectively.

The formula for $SIM_{part}$ is the same as that of $SIM$, except that the meaning of $\mathcal{M}_{gt}^{2D}$ differs. For an object with a handle, its graspable part is actually the entire handle. Thus, any point falling within the handle region after affordance argmax should be considered valid. Therefore, in $SIM_{part}$, we use the binary mask of the handle region as the ground truth for calculation.

\subsection{Dataset Taxonomy}

\begin{figure}[htbp]
    \centering
    \includegraphics[width=1.0\linewidth]{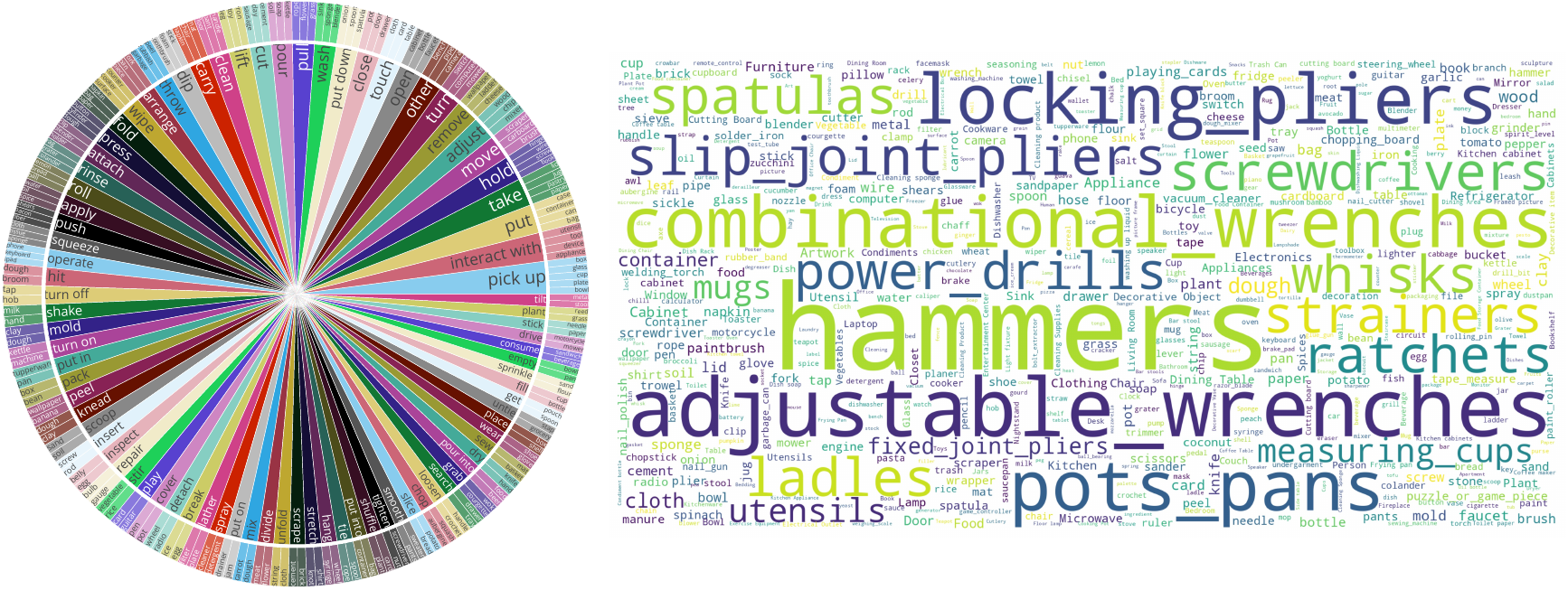}
    \caption{\textbf{Left:} The distribution of primary actions and related object categories in HOVA-500K ($>$100 data samples). \textbf{Right:} Word cloud of primary object categories in HOVA-500K. }
    \label{fig:data_vis_appendix}
\end{figure}

\begin{figure}[htbp]
    \centering
    \includegraphics[width=1\linewidth]{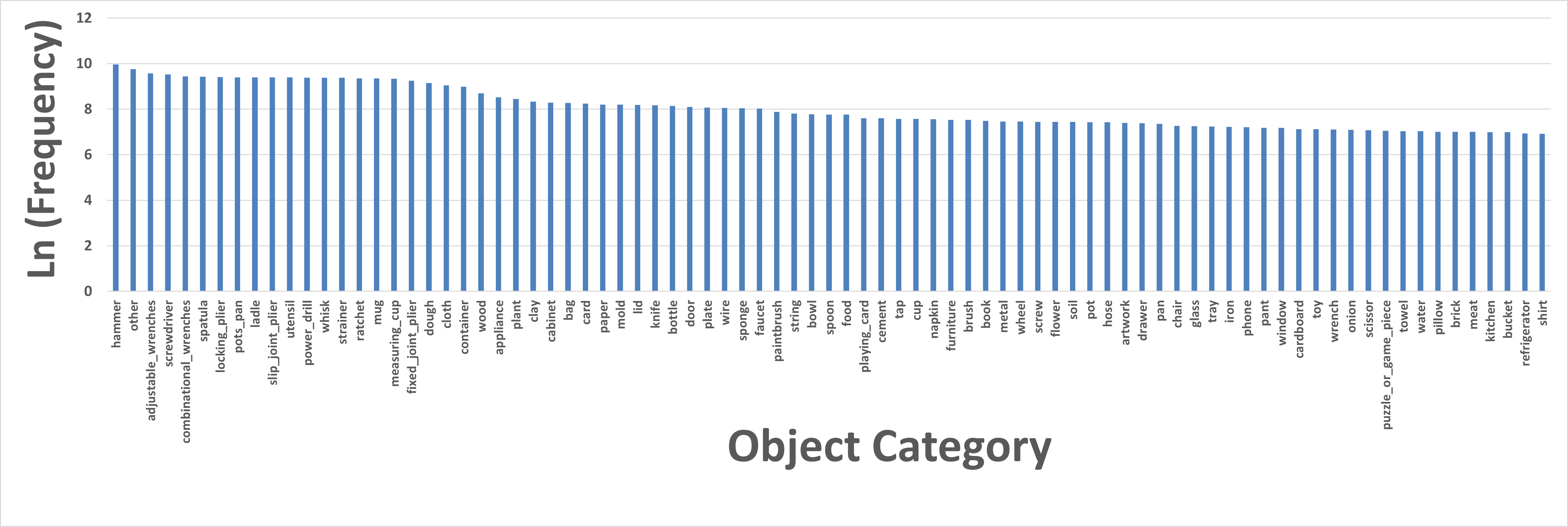}
    \caption{The distribution of primary object categories in HOVA-500K ($>$1000 data samples)}
    \label{fig:obj_ft_1000}
\end{figure}

\begin{figure}[htbp]
    \centering
    \includegraphics[width=1\linewidth]{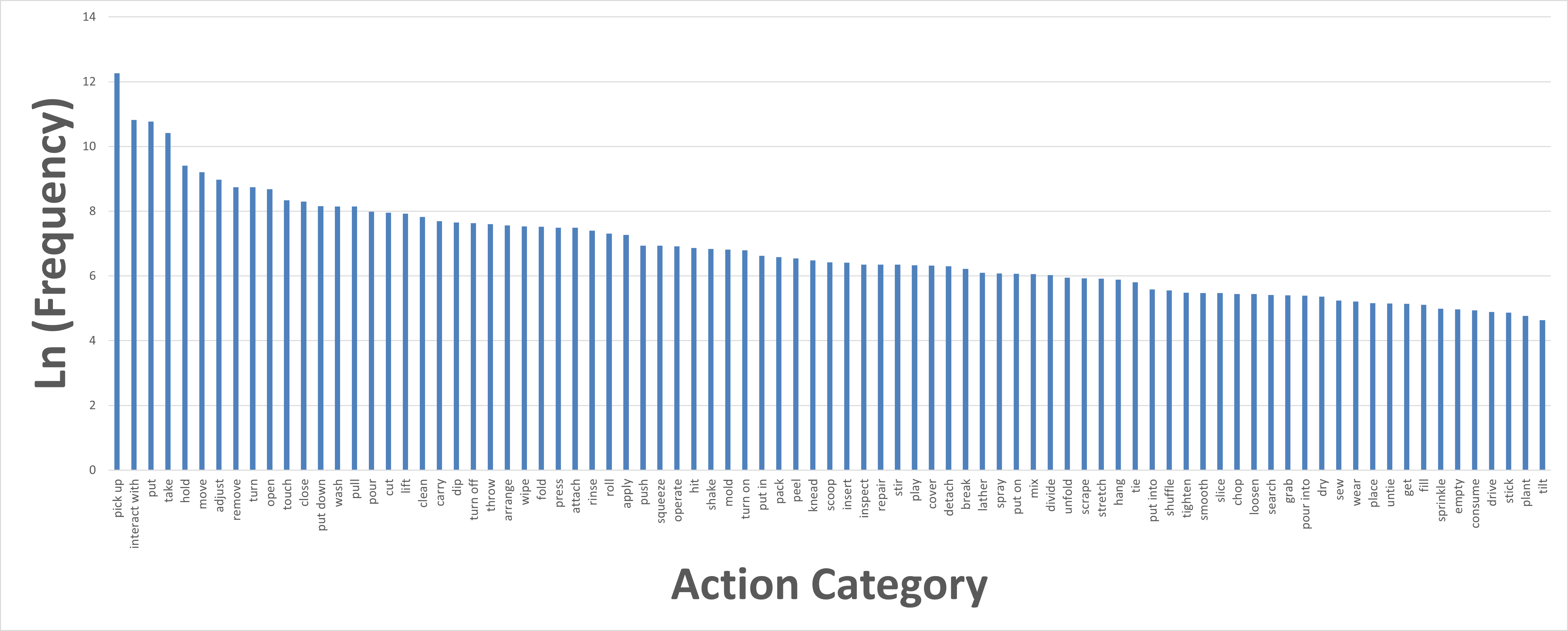}
    \caption{The distribution of primary action categories in HOVA-500K ($>$100 data samples)}
    \label{fig:action_ft_100}
\end{figure}

Our dataset consists of 1,726 object categories and 675 verb categories. The object categories can be referenced in the right panel of Figure \ref{fig:data_vis_appendix}, with examples including \textit{hammer, screwdriver, adjustable wrenches, combinational wrenches, spatula, etc}. Some verb categories are shown in the left panel of Figure \ref{fig:data_vis_appendix}, with examples such as \textit{pick up, interact with, put, take, hold, move, etc}. Figures \ref{fig:obj_ft_1000} and \ref{fig:action_ft_100} display the logarithmic frequency distribution histograms for object categories with more than 1,000 instances and verb categories with more than 100 instances, respectively.

\section{GLOVER++ Model}
\subsection{Model Pipeline}
\label{sec:model_appendix}
\begin{figure}[htbp]
    \centering
    \includegraphics[width=1.0\linewidth]{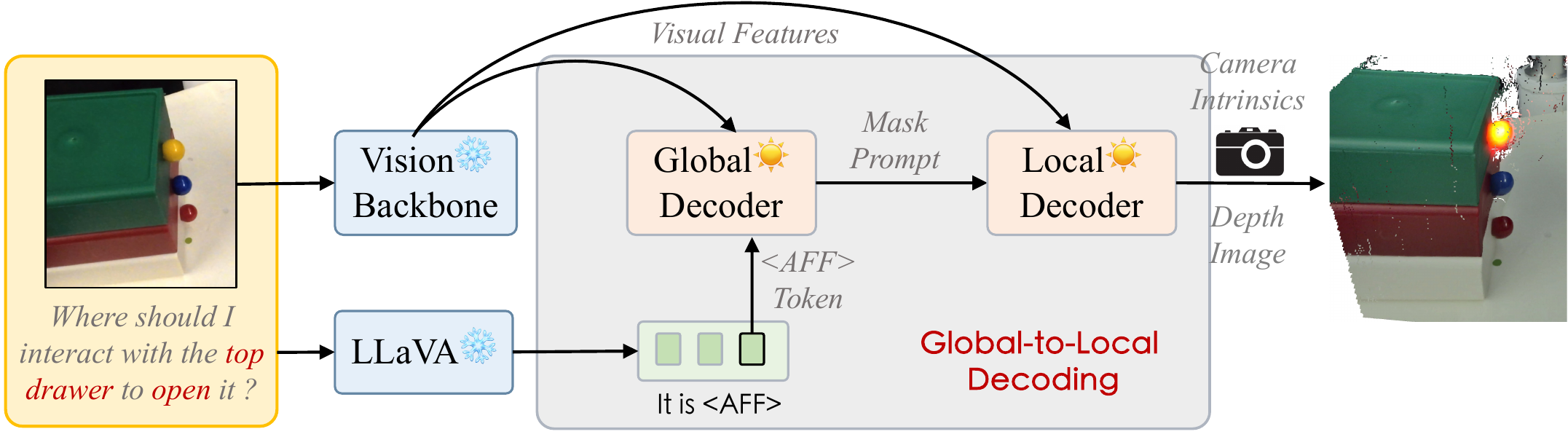}
    \caption{The pipeline of GLOVER++.}
    \label{fig:glover++_pipeline}
\end{figure}

We show the detailed pipeline of GLOVER++ in Fig.~\ref{fig:glover++_pipeline}. We utilize the pretrained vision backbone and LLaVA model from the LISA++~\cite{yang2023lisa++} to inherit the open-vocabulary reasoning knowledge from it. The vision backbone and LLaVA model are frozen during the training. We follow the SAM~\cite{sam} to design the decoders, and both the global and local decoder comprises two layers of bi-directional Transformer layers. 


The first global decoder processes the \texttt{<AFF>} token to generate a semantic-aware logits map, where the token aggregates vision-language features from the input to encode high-level image semantics. While this global understanding captures contextual relationships, it inevitably introduces background noise—irrelevant regions activated by broad semantic correlations (e.g., "cut" may falsely highlight all sharp objects). Such noise conflicts with the localized nature of affordance learning, which demands precise spatial grounding of action-relevant object parts.

\begin{figure}[htbp]
    \centering
    \includegraphics[width=1.0\linewidth]{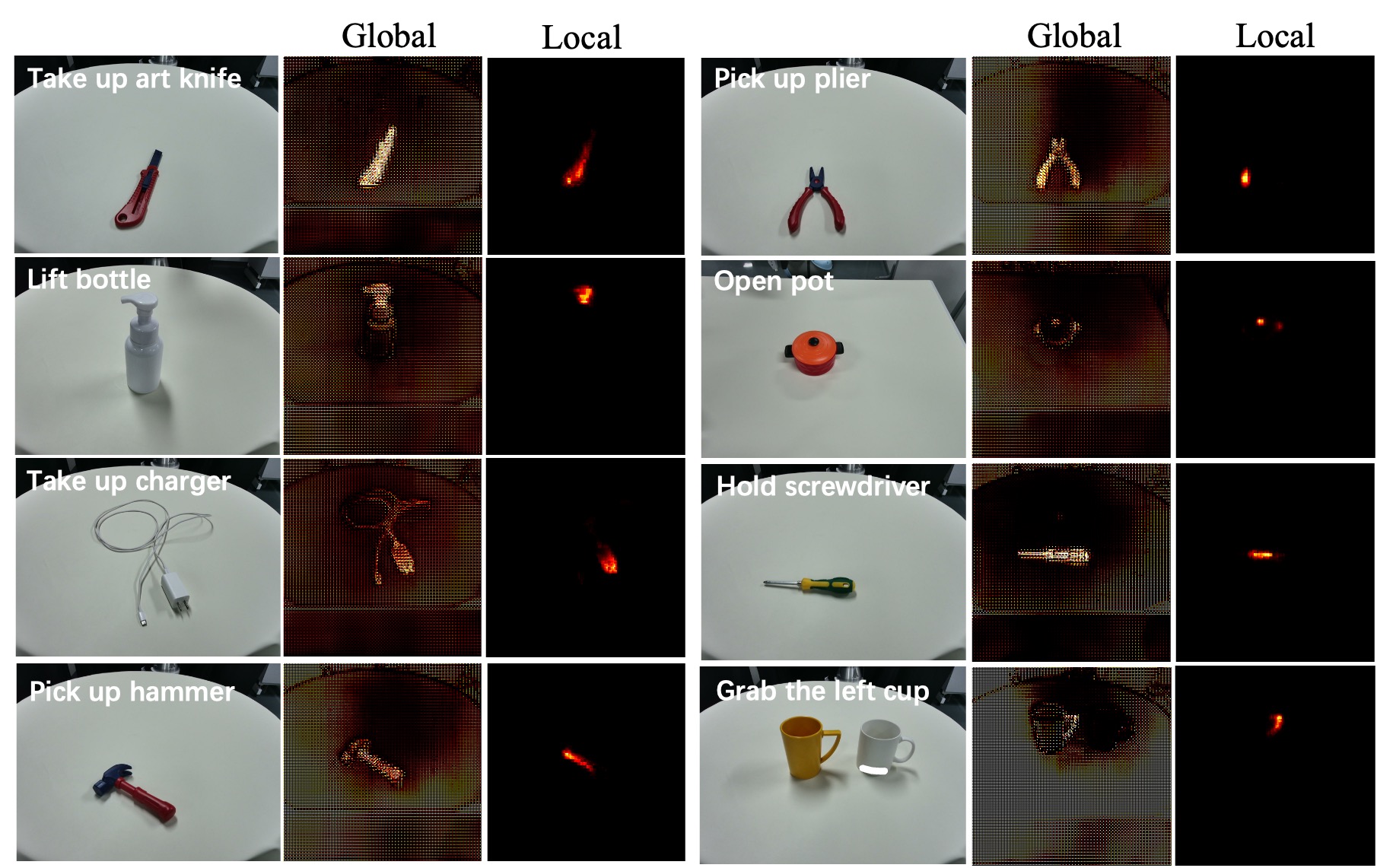}
    \caption{More visualization of the decoded features by the global and local decoder }
    \label{fig:global-to-local-appendix}
\end{figure}

To resolve this discrepancy, we propose a cascaded local decoder that refines the global decoder’s output. The local decoder treats the initial semantic logits map as a mask prompt, dynamically focusing on regions where language instructions align with local object geometries. This hierarchical design synergizes global semantic priors with local affordance specificity: the global decoder provides object-level awareness, while the local decoder resolves part-level actionable regions. We show the effectiveness of this global-to-local mechanism in Fig.~\ref{fig:global-to-local-appendix}. Also, in this process, we hope to preserve the world knowledge and open-vocabulary reasoning capability of the LLaVA in the first global decoding, thus distributing a relatively small learning rate to update the global decoder. For the local decoder, a large learning rate is set to ensure the thorough learning of the affordance representation.

Compared with GLOVER~\cite{ma2024glover}, GLOVER++ introduces negligible additional trainable parameters, as shown in Table~\ref{tab:param}, while achieving much better affordance reasoning performance. 

\begin{table}[htbp]
  \centering
  \renewcommand{\arraystretch}{1.3}
  \resizebox{0.8\textwidth}{!}{
   \begin{tabular}{ccccccc}
      \toprule
      \textbf{Methods} & \textbf{Trainable \#param} & \textbf{Ratio} & \textbf{KLD} $\downarrow$ & \textbf{SIM} $\uparrow$ & $\textbf{SIM}_{part}$ $\uparrow$ & \textbf{NSS} $\uparrow$ \\
      \hline
    GLOVER~\cite{ma2024glover} & 4.1M & 0.0526\% & 7.441 & 0.025 & 0.206 & 0.900 \\
      \rowcolor{gray!40}
      \textbf{GLOVER++} & 8.1M & 0.1050\% & \textbf{3.411} & \textbf{0.120} & \textbf{0.506} & \textbf{5.151} \\ 
      \bottomrule
   \end{tabular}
   }
    \vspace{5pt}
   \caption{Trainable parameters and their ratio to the total parameters. Our model surpasses GLOVER in performance with only a marginal increase in trainable parameters.}
   \label{tab:param}
\end{table}

\subsection{Training Objective}
\label{sec:Training Objective}

The final training objective consists of two parts: the sigmoid focal loss and the Kullback-Leibler Divergence (KLD) loss. Here, $\alpha, \gamma$ denotes the focusing and balancing parameters in the focal loss, respectively. Additionally, $\alpha_t, p_t$ represent the soft versions of $\alpha, p$ in focal loss formulation. $g_i, p_i$ correspond to the ground truth and predicted affordance value, respectively. 
\begin{equation}
    \mathcal{L}=-\lambda_{Sigmoid}\sum_{i}{\alpha_t(1-p_t)^\gamma\mathcal{L}_{CE}+\lambda_{KL}\sum_{i}g_i\log\frac{g_i}{p_i}}
\end{equation}
\begin{equation}
    \mathcal{L}_{CE}=-g_i\log{p_i}
\end{equation}
\begin{equation}
    \alpha_t=\alpha g_i+(1-\alpha)(1-g_i)
\end{equation}
\begin{equation}
    p_t=p_ig_i+(1-p_i)(1-g_i)
\end{equation}

\section{Experiments}
\subsection{GLOVER++ Training Details.}
\label{sec:training_details}
Our training is conducted on 8 NVIDIA 48G A6000 GPUs for 10 epochs. The training scripts are based on deepspeed engine\cite{rasley2020deepspeed}. We employ AdamW\cite{loshchilov2017decoupled} optimizer($\beta_1=0.9,\beta_2=0.95$) with a weight decay of 0.0005. The learning rates for the mask decoder and affordance decoder are set to 5e-5 and 5e-4, respectively. We use WarmupDecayLR as the learning rate scheduler, with 188 warmup steps. The KL Loss and Focal Loss are both weighted at 0.1. Additionally, the batch size per GPU is configured as 32. The mask decoder is initialized with pretrained weights from LISA++\cite{yang2023lisa++}, while the affordance decoder adopts weights from SAM \cite{sam}.

\subsection{Imitation Learning in RLBench}
\label{sec:IL_appendix}

\begin{figure}[htbp]
    \centering
    \includegraphics[width=1.0\linewidth]{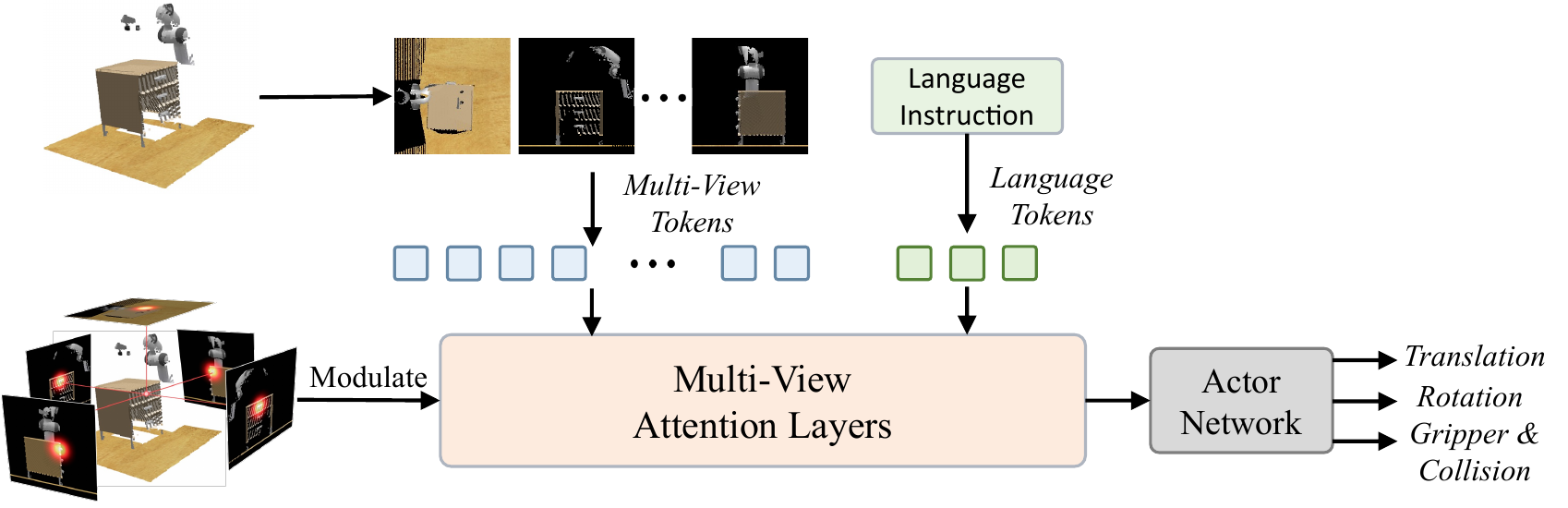}
    \caption{The pipeline of using actionable affordance as prior to guide the attention of multi-task language-guided manipulation.}
    \label{fig:rlbench_pipeline}
\end{figure}
 
We aim to demonstrate that explicit affordance representation can effectively guide imitation learning networks to focus on action-critical regions in visual inputs. 
We adopt RLBench~\cite{james2020rlbench} as a testbed to demonstrate. RLBench is a high-fidelity simulated environment
built on PyRep (PyBullet wrapper)~\cite{james2019pyrep} with stable physics.  

Our training closely follows RVT~\cite{goyal2023rvt}, utilizing cube-view re-rendered images generated from 3D point clouds~\cite{ma2024contrastive}. To ensure efficiency, the replay buffer of extracted keyframes is adopted to train the agent rather than all frames from episodes. 
For data augmentation, we adopt PerAct's~\cite{peract2022arxiv} approach, applying random perturbation of translations within $\pm0.125m$ and rotations along $z\mbox{-}$axis within $\pm45^{\circ}$. We train the RVT and RVT-AFF for 5 epochs with a batch size of 24 and a learning rate of 1e-4. For the RVT2 and RVT2-AFF, we train the models for 100 epochs with a batch size of 24 and a learning rate of 1.25e-5. 

\begin{table}[]
\centering
\resizebox{1.0\linewidth}{!}{
\begin{tabular}{llcc}
\toprule
 Task & Language Template & \# of Variations  &Avg. Keyframes\\ 
\midrule
open drawer &  ``open the \blank drawer" & 3 &3.0 \\
slide block & ``slide the \blank block to target" & 4  &4.7\\
sweep to dustpan & ``sweep dirt to the \blank dustpan" & 2  &4.6\\
meat off grill & ``take the \blank off the grill" & 2 &5.0\\
turn tap & ``turn \blank tap" & 2 &2.0 \\
put in drawer & ``put the item in the \blank drawer" & 3 &12.0\\
close jar & ``close the \blank jar" & 20 &6.0\\
drag stick & ``use the stick to drag the cube onto the \blank target" & 20 &6.0\\
stack blocks & ``stack \blank \blank blocks'' & 60 &14.6\\
screw bulb  & ``screw in the \blank light bulb” & 20 &7.0\\
put in safe  & ``put the money away in the safe on the \blank shelf'' & 3 &5.0\\
place wine  & ``stack the wine bottle to the \blank of the rack'' & 3 &5.0\\
put in cupboard & ``put the \blank in the cupboard'' & 9 &5.0\\
sort shape & ``put the \blank in the shape sorter'' & 5 &5.0\\
push buttons & ``push the \blank button, [then the \blank button]'' & 50 &3.8\\
insert peg  & ``put the \blank peg in the spoke'' & 20 &5.0\\
stack cups  & ``stack the other cups on top of the \blank cup'' & 20 &10.0\\
place cups  & ``place \blank cups on the cup holder''  & 3 &11.5\\
\bottomrule
\end{tabular}}
\vspace{2mm}
\caption{Tasks we used in RLBench. }
 \label{tab:rlbench_tasks}
\end{table}

\subsection{Ablations}
\label{sec:ablation_appendix}
\begin{figure}[htbp]
    \centering
    \includegraphics[width=0.7\linewidth]{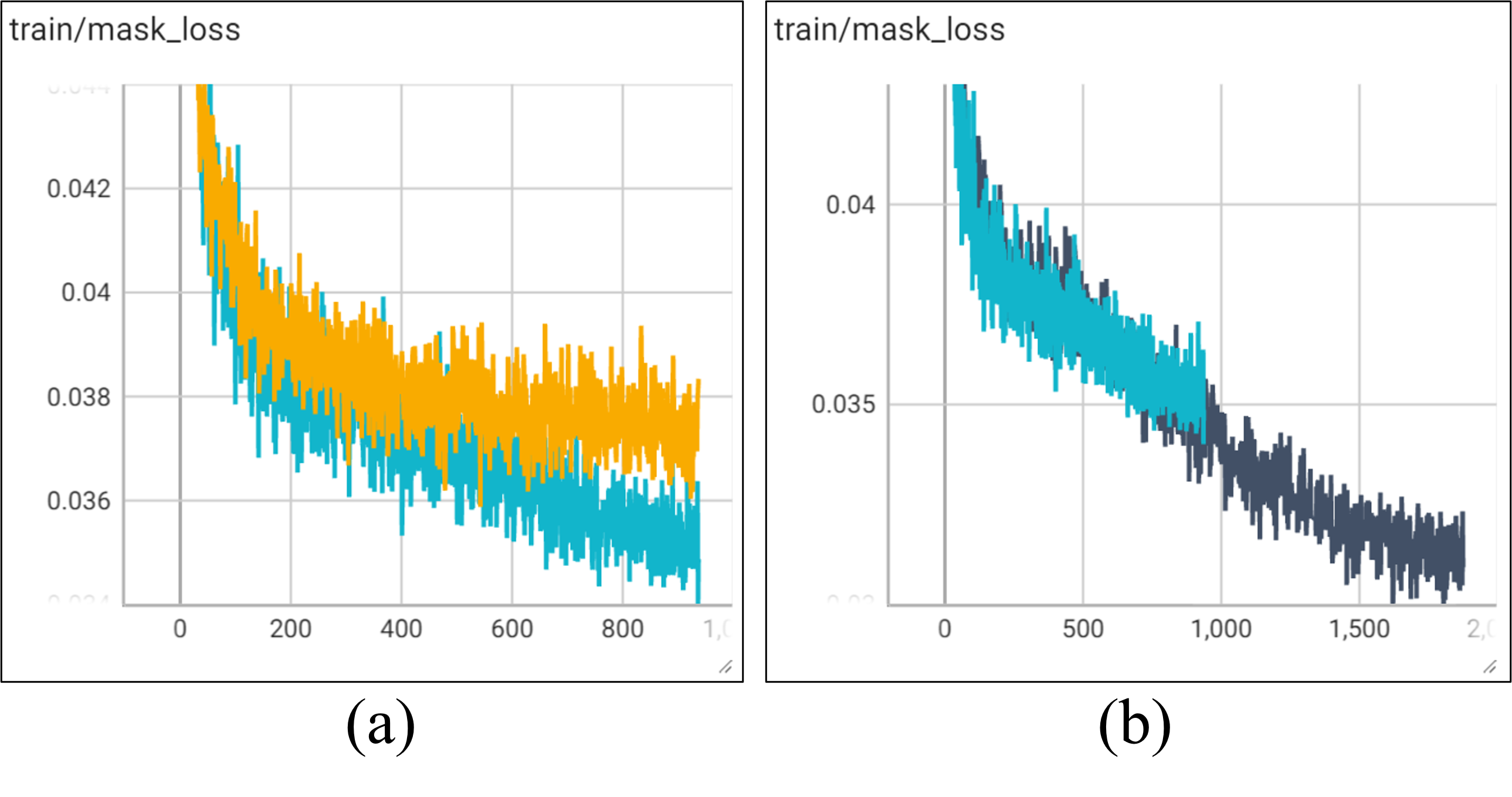}
    \vspace{-5pt}
    \caption{\textbf{The loss curves of training}. (\textbf{a}): The yellow and blue curve represents model w/o and w/ global-to-local decoding module, respectively. (\textbf{b}):The blue loss curve reflects a 5-epoch scheme of training, while the black one reflects a 10-epoch one.}
    \label{fig:loss_compare}
\end{figure}

We show more visualization of ablative studies to further demonstrate the effectiveness of the proposed components. 
The loss curves in Fig.~\ref{fig:loss_compare} (a) show that global-to-local decoding leads to effective convergence of the total loss function, which conforms to better performance in Table~\ref{tab:affordance}. Also, the  Fig.~\ref{fig:loss_compare} (b) reveals the advantage of extending the training scheme in the GLOVER++'s fine-tuning.

\begin{figure}[htbp]
\vspace{-15pt}
    \centering
    \includegraphics[width=0.7\linewidth]{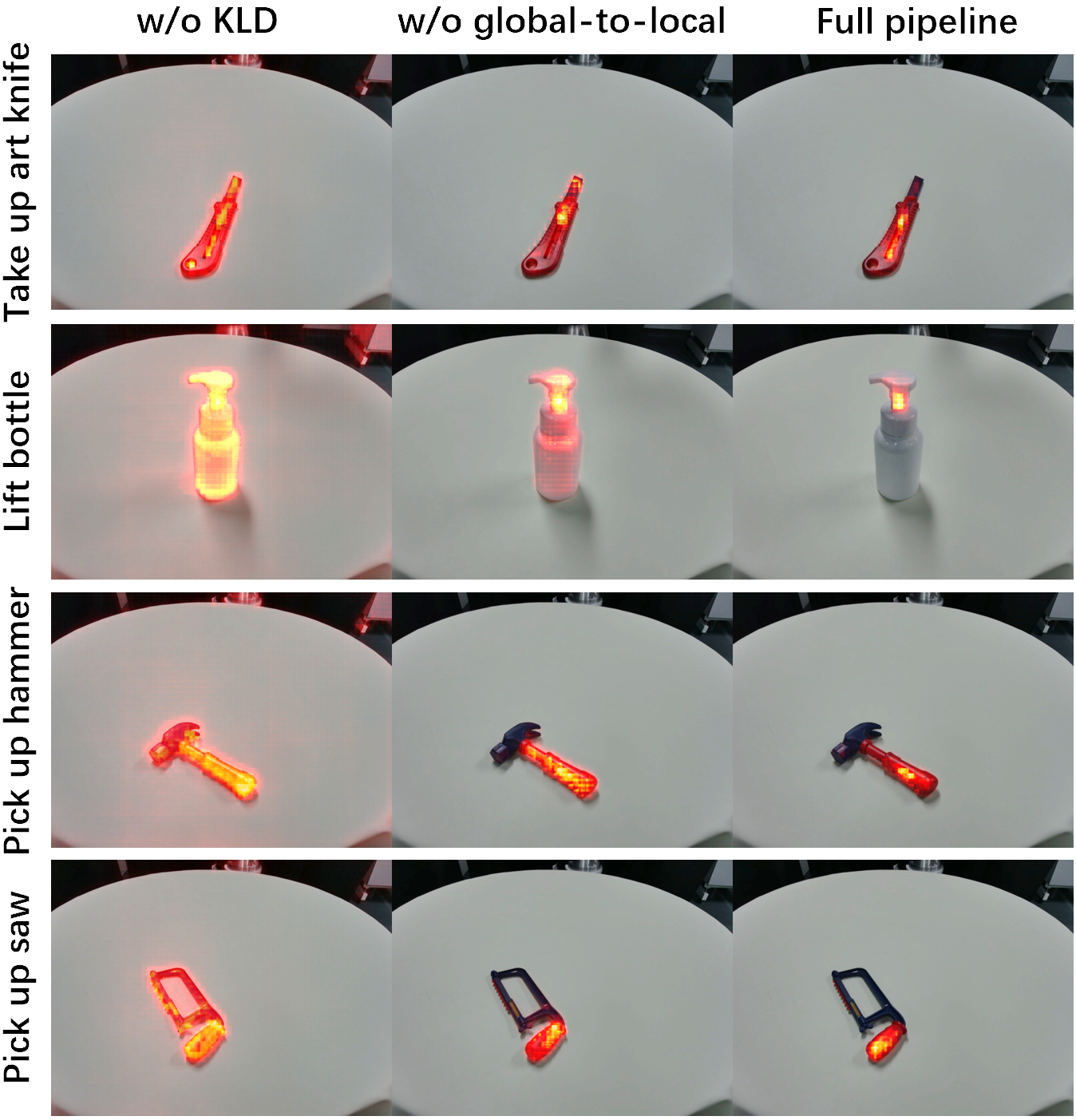}
    \caption{Visualization of the effectiveness of GLOVER++'s components in affordance reasoning.}
    \label{fig:ablation_compare_vis}
\end{figure}

The Fig.~\ref{fig:ablation_compare_vis} illustrates the affordance reasoning visualization for ablative comparisons. Clearly, KLD loss and global-to-local decoding optimize local affordance representation learning for GLOVER++.

\subsection{More Comparisons with Qwen-2.5}
\begin{figure}[htbp]
    \centering
    \includegraphics[width=0.8\linewidth]{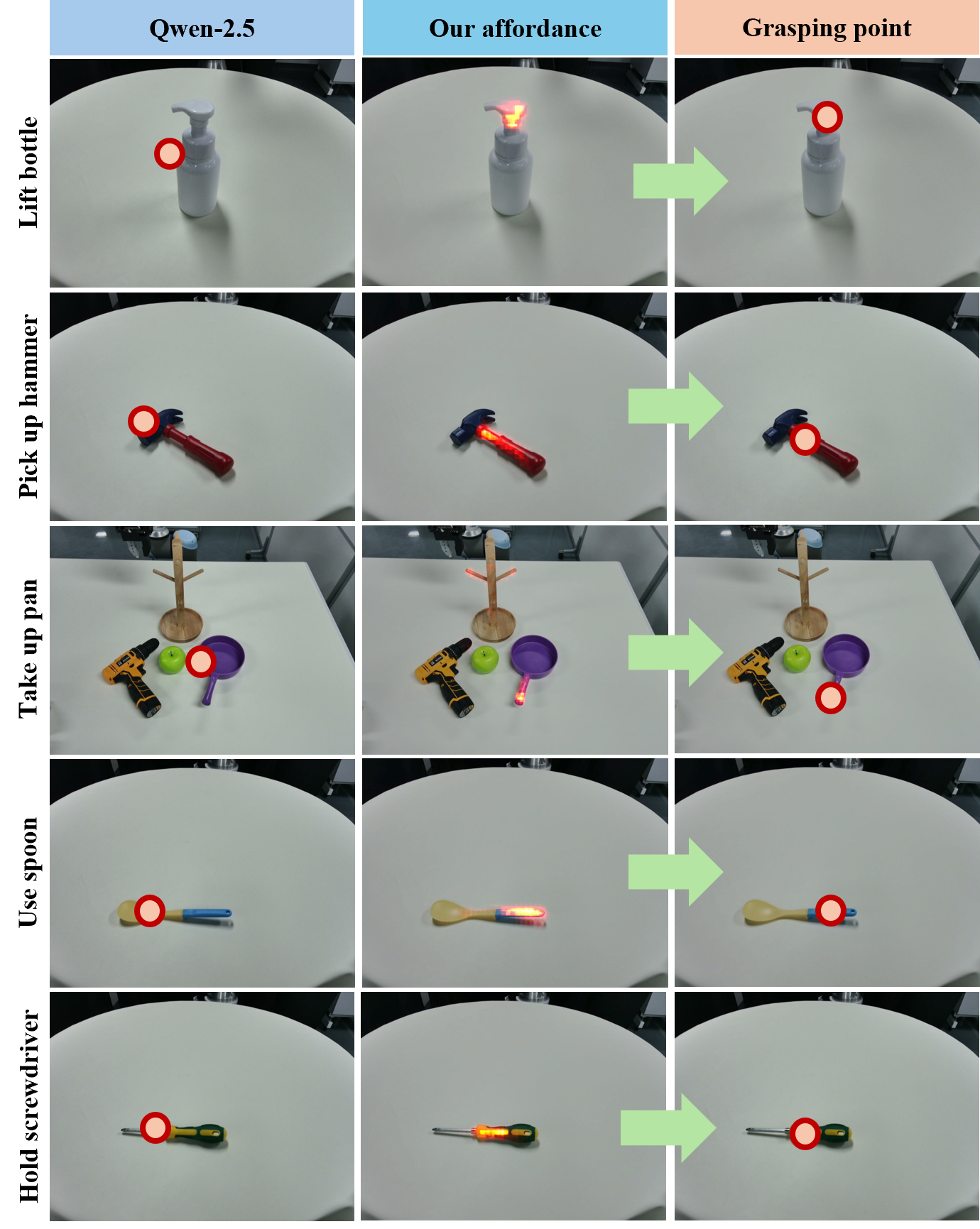}
    \caption{More visualization of comparison with Qwen-2.5-VL~\cite{yang2024qwen2}. We show the inferred grasping point of GLOVER++ by argmaxing the affordance regions.}
    \label{fig:compare-qwen-appendix}
\end{figure}

We show more comparisons between GLOVER++ and Qwen-2.5-VL-7B~\cite{yang2024qwen2} model in Fig.~\ref{fig:compare-qwen-appendix}. Compared to Qwen-2.5-VL, GLOVER++ demonstrates superior capability in generating task-compliant grasp points by explicitly modeling action-object semantics in language instructions.
While Qwen-2.5-VL primarily relies on visual-language alignment for object localization, it often fails to disambiguate action-specific affordances to infer reasonable grasping regions.

\subsection{Real-world Experiments Setting}
\begin{figure}
    \centering
    \includegraphics[width=0.9\linewidth]{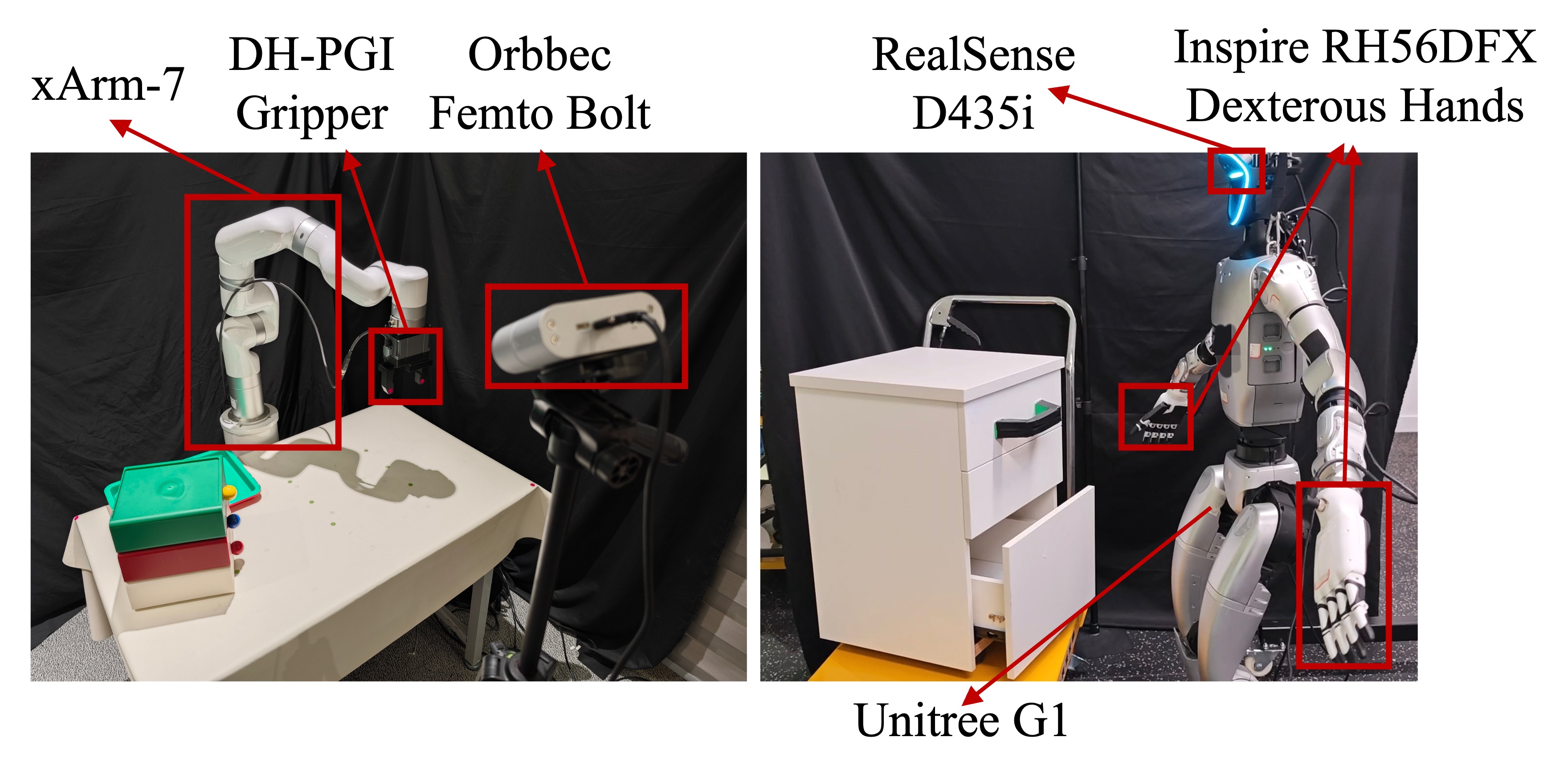}
    \caption{The real-world experiments settings.}
    \label{fig:real_world_setting}
\end{figure}

In the real-world experiments, we adopt two systems for manipulation tasks as Fig.~\ref{fig:real_world_setting} shows. 
The first system is based on a 7-DoF UFACTORY xAarm 7, equipped with DH-PGI gripper. We utilize an Orbbec Femto Bolt RGB-D camera for visual observations, with the image size of $1280\times 960$ by default. The inverse kinematics (IK) to resolve the trajectory planning. 

\begin{figure}
    \centering
    \includegraphics[width=0.5\linewidth]{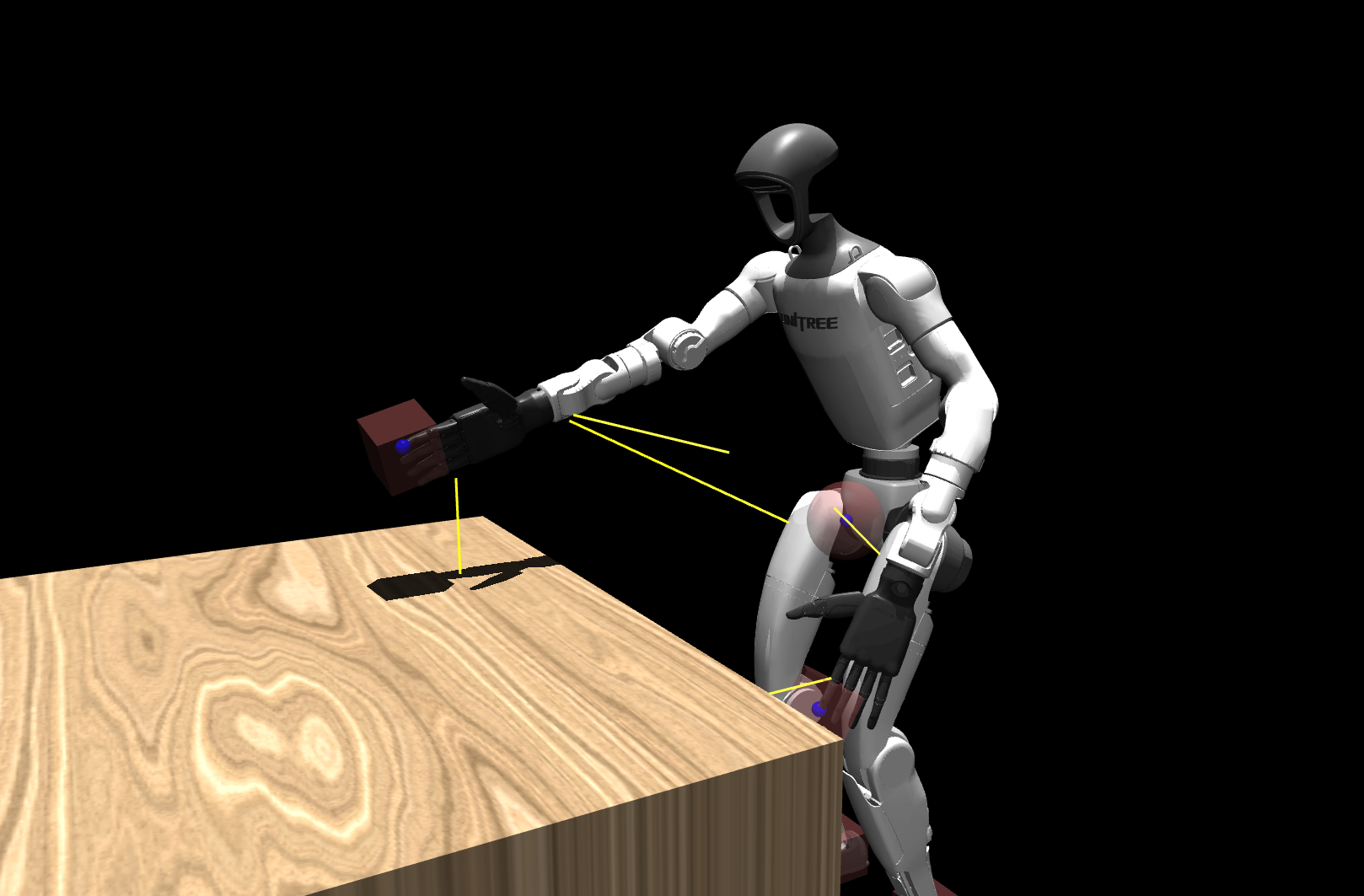}
    \caption{The illustration of the obstacle-avoidance IK we use to avoid self-collision.}
    \label{fig:humanoid_avoid}
\end{figure}

On the other hand, we leverage a Unitree G1 humanoid robot with two Inspire dexterous hands RH56DFX to construct a system for dexterous and bimanual grasping. We use the original head-mounted Intel394
RealSense D435i in Unitree G1 to capture RGB-D images with the size of $640\times 480$. For the motion planning, we use the obstacle-avoidance IK to avoid the self-collision as shown in Fig.~\ref{fig:humanoid_avoid}.

\subsection{Failure Case}
\label{sec:failure_case}
We illustrate the failure cases in the aspect of both affordance reasoning and real-world experiments.  

\begin{figure}[htbp]
    \centering
    \includegraphics[width=0.8\linewidth]{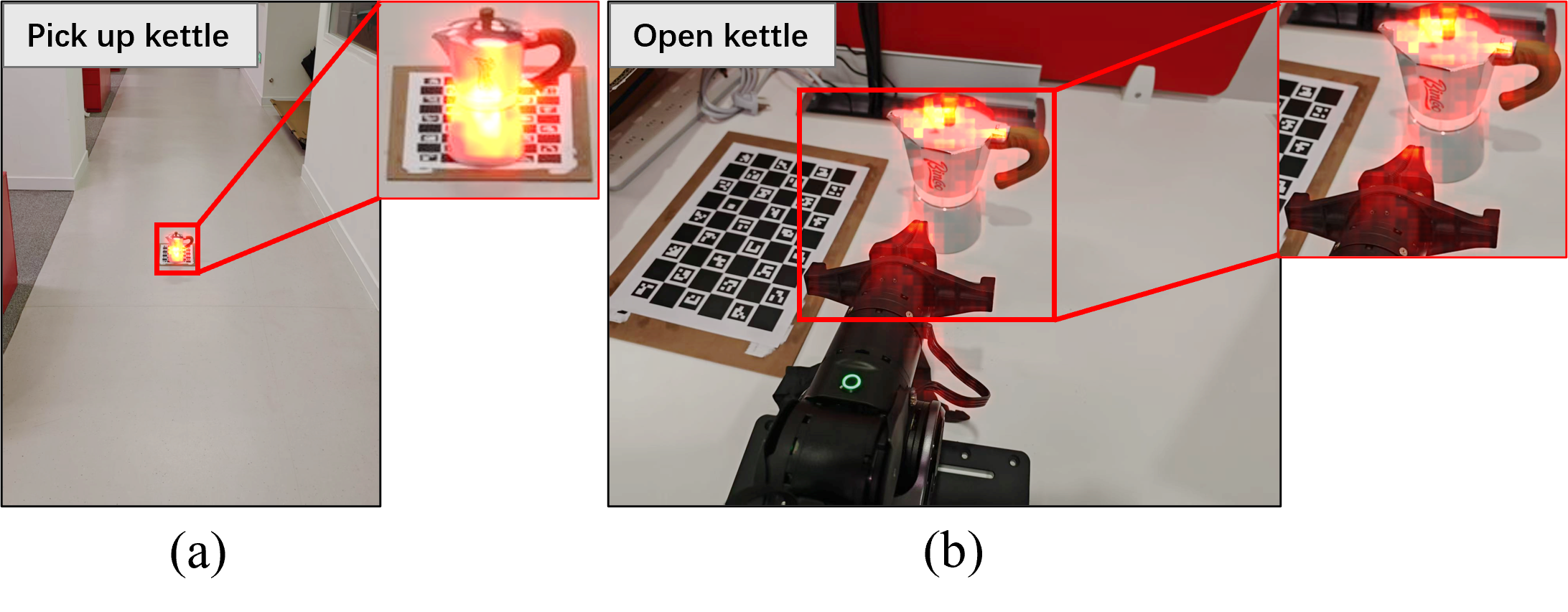}
    \vspace{-10pt}
    \caption{The failure cases of affordance reasoning, including: (a) distance viewpoint leads to inaccurate affordable regions, (b) failure to distinguish between overlapping objects sometimes, and results in background noise. }
    \label{fig:failure_case_affordance}
\end{figure}

\begin{figure}[htbp]
\vspace{-10pt}
    \centering
    \includegraphics[width=1.0\linewidth]{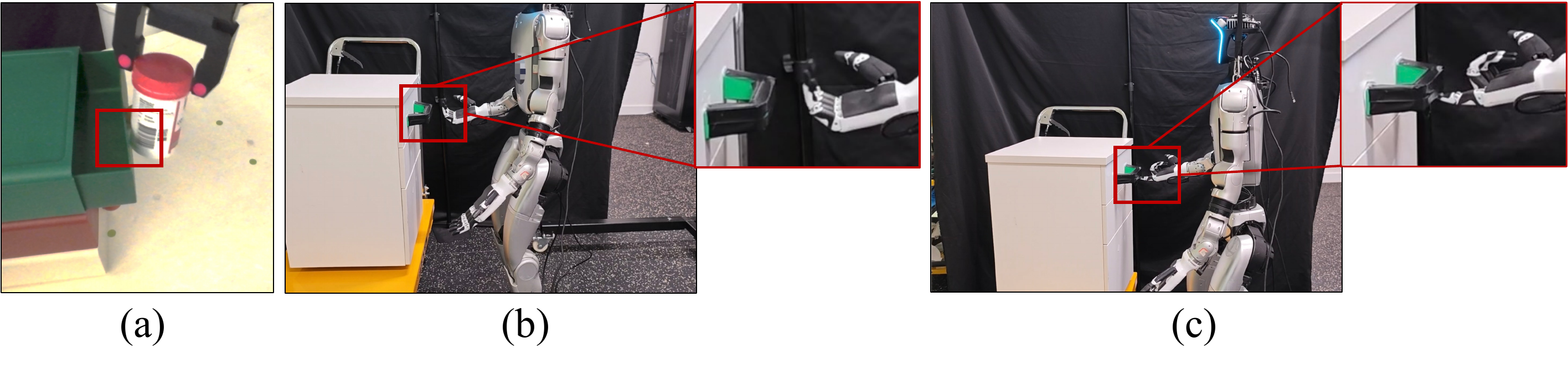}
    \caption{Failure cases in real-world experiments, including (a) collision problem, (b) z-axis inaccuracy, (c) imperfect grasping pose. }
    \label{fig:failure_case_real}
\end{figure}

\noindent \textbf{Affordance Reasoning.} 
The failure cases in the affordance reasoning includes two aspects as far as we know. First, when the viewpoint is excessively distant, GLOVER++ struggles to infer precise grasping regions and can only predict coarse object-level locations, as shown in Fig.~\ref{fig:failure_case_affordance} (a). Second, since the affordance reasoning is performed in 2D space, GLOVER++ sometimes struggles to distinguish between overlapping objects at the pixel level, leading to noisy probability maps, although the highest-probability grasp points remain correct. The Fig.~\ref{fig:failure_case_affordance} (b) shows the circumstance.

\noindent \textbf{Real-world Experiments.} 
In the real-world experiments, the failure cases result from three primary reasons: (1) The collision caused by the imperfect rollout planning. (2) The projected affordable points may exhibit z-axis distance inaccuracies with one RGB-D camera, leading to the grasping failure. (3) The imperfect grasping pose of high-DoF dexterous hands leads to task failure. We show the above three cases in the Fig.~\ref{fig:failure_case_real}.


\begin{figure}[htbp]
    \centering
    \includegraphics[width=1.0\linewidth]{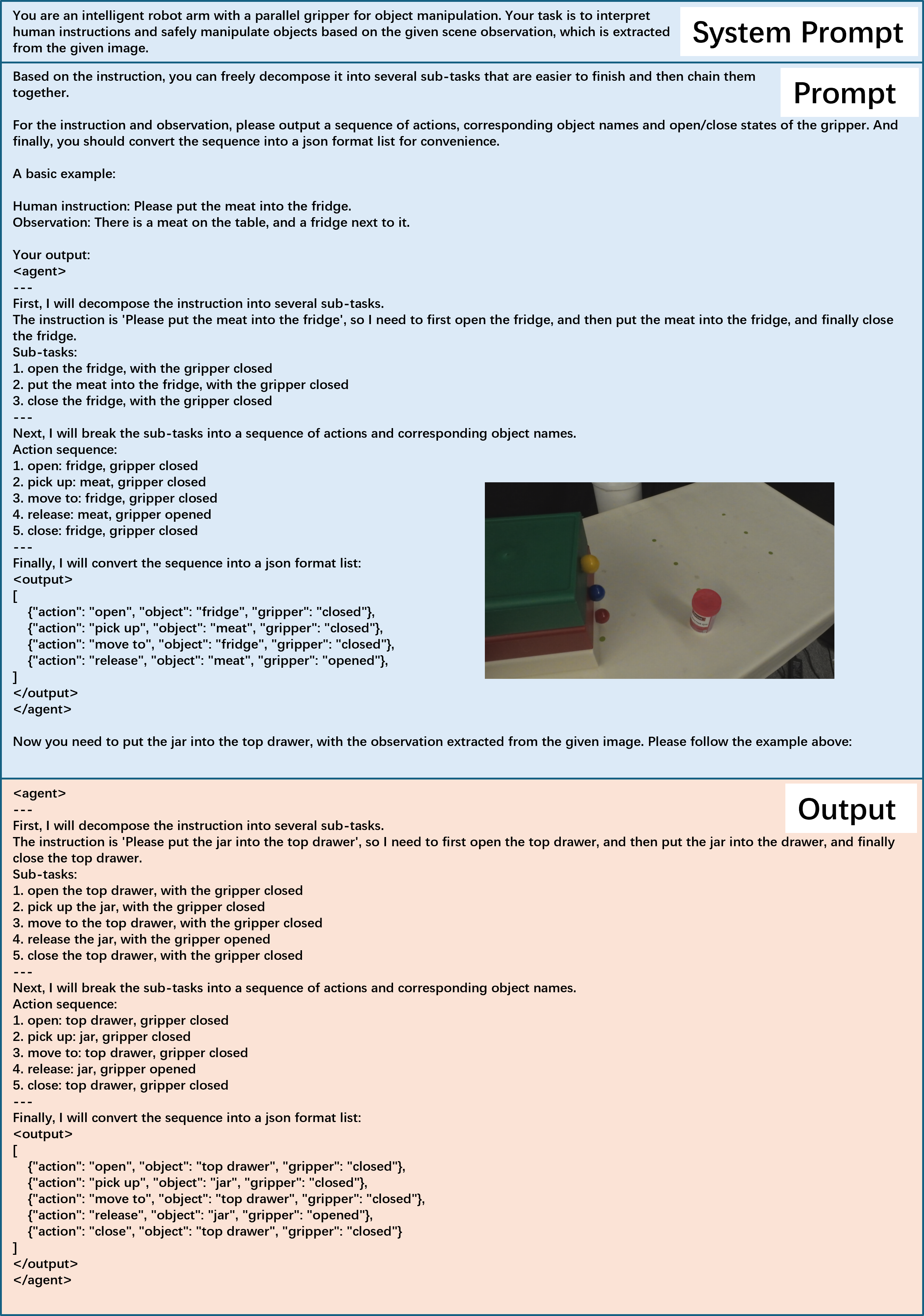}
    \caption{An example of prompting Qwen-2.5-VL-7B to decompose the long-horizon task for GLOVER++.}
    \label{fig:qwen_planner}
\end{figure}

\end{document}